\documentclass[10pt,twocolumn,letterpaper]{article}

\usepackage{cvpr}
\usepackage{times}
\usepackage{epsfig}
\usepackage{graphicx}
\usepackage{amsmath}
\usepackage{amssymb}
\usepackage{bbold}
\usepackage{dsfont}

\usepackage{enumerate}
\usepackage[toc,page]{appendix}
\usepackage[font=small,labelfont=bf]{caption}

\usepackage{xcolor}
\usepackage{amsfonts}

\usepackage{caption}
\usepackage{subcaption}
\usepackage{bm}
\usepackage{isomath}

\usepackage[numbers]{natbib}

\usepackage[pagebackref=true,breaklinks=true,letterpaper=true,colorlinks,bookmarks=false]{hyperref}
\usepackage{footmisc}
\usepackage{stmaryrd}

\usepackage{fixltx2e}
\usepackage{dblfloatfix}
\usepackage{pbox}
\usepackage{capt-of}

\usepackage[normalem]{ulem}
\usepackage{multirow}

\usepackage{colortbl}

\usepackage{mathtools}
\usepackage{array}

\usepackage{amsmath}
\usepackage{amssymb}
\usepackage{amsthm}
\usepackage{indentfirst}

\usepackage{arydshln}
\usepackage{multirow}
\usepackage{multicol}

\usepackage{url}
\usepackage{enumitem}

\newcommand*{\Scale}[2][4]{\scalebox{#1}{$#2$}}

\makeatletter
\@namedef{ver@everyshi.sty}{}
\makeatother
\usepackage{pgf}

\renewcommand\cdots{...}

\newcommand{\mX}{\mathbf{X}}

\newcommand{\mbr}[1]{\mathbb{R}^{#1}}

\newcommand{\vpsi}{\boldsymbol{\psi}}

\def\eg{\emph{e.g.}}

\newcommand{\mPhi}{\boldsymbol{\Phi}}

\newcommand{\stkout}[1]{{\ifmmode\text{\sout{\ensuremath{#1}}}\else\sout{#1}\fi}}

\newcommand{\comment}[1]{}

\usepackage[pagebackref=true,breaklinks=true,letterpaper=true,colorlinks,bookmarks=false]{hyperref}

\cvprfinalcopy % *** Uncomment this line for the final submission

\def\cvprPaperID{1937} % *** Enter the CVPR Paper ID here

\ifcvprfinal\fi
\begin{document}

%%%%%%%%% TITLE
\title{Few-shot Learning via Saliency-guided Hallucination of Samples}

\author{Hongguang Zhang\textsuperscript{1,2}\qquad Jing Zhang\textsuperscript{1,2}\qquad Piotr Koniusz\textsuperscript{2,1}\\
$^1$Australian National University, \quad$^2$Data61/CSIRO \\
firstname.lastname@\{anu.edu.au\textsuperscript{1}, data61.csiro.au\textsuperscript{2}\}
}
\maketitle
\thispagestyle{empty}
\vspace{-0.5cm}

%%%%%%%%% ABSTRACT
\begin{abstract}
Learning new concepts from a few of  samples is a standard challenge in computer vision. The main directions to improve the learning ability of few-shot training models include (i) a robust similarity learning and (ii) generating or hallucinating additional data from the limited existing samples. In this paper, we follow the latter direction and present a novel data hallucination model. Currently, most datapoint generators contain a specialized network (\ie, GAN) tasked with hallucinating new datapoints, thus requiring large numbers of annotated data for their training in the first place. In this paper, we propose a novel less-costly hallucination method for few-shot learning which utilizes saliency maps.
To this end, we employ a saliency network to obtain the foregrounds and backgrounds of available image samples and feed the resulting maps into a two-stream network to hallucinate datapoints directly in the feature space from viable foreground-background combinations. To the best of our knowledge, we are the first to leverage saliency maps for such a task and we demonstrate their usefulness in hallucinating additional datapoints for few-shot learning. Our proposed network achieves the state of the art on publicly available datasets.
\end{abstract}

%%%%%%%%% BODY TEXT
\vspace{-0.5cm}
\section{Introduction}
\label{sec:intro}
Convolutional Neural Networks (CNN) have demonstrated their usefulness in numerous computer vision tasks \eg, image classification and scene recognition. However, training CNNs on these tasks requires large numbers of labeled data. In contrast to CNNs, human ability to learn novel concepts from a few of samples remains unrivalled. Inspired by this observation, researchers \cite{fei2006one} proposed the one- and few-shot learning tasks with the goal of training algorithms with low numbers of datapoints.

Recently, the concept of learning relations with deep learning has been explored in several papers \cite{vinyals2016matching,snell2017prototypical,sung2017learning,NIPS2017_7082} which can be viewed as a variant of metric learning \cite{metric_old,kissme,Mehrtash_CVPR_2018} adapted to the few-shot learning scenario. In these works, a neural network extracts convolutional descriptors, and another learning mechanism (\eg, a relation network) captures relationship between descriptors. Most papers in this category propose improvements to relationship modeling for the purpose of similarity learning. In contrast, \cite{hariharan2017low} employs a separate Multilayer Perceptron (MLP)  to  hallucinate additional image descriptors by modeling foreground-background relationships in feature space to obtain implicitly augmented new samples. To train the feature generator, MLP uses manually labelled features clustered into 100 clusters, which highlights the need for extra labelling. Another approach \cite{wang2018low} generates data in a meta-learning scenario, which means the network has to be pre-trained on several datasets, thus increasing the cost of training.

\begin{figure}[t]
    \vspace{-0.4cm}
    \centering
    \includegraphics[width=\linewidth]{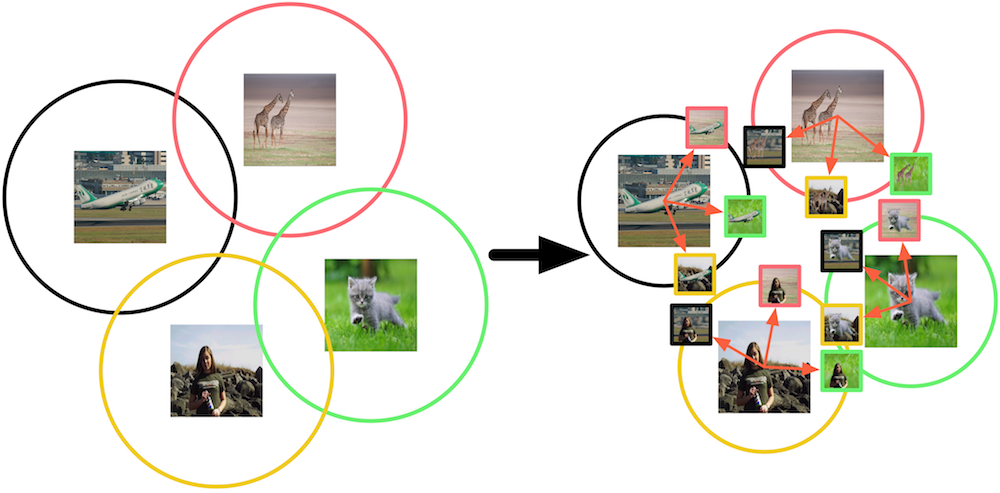}
    \caption{Illustration of saliency-based data generation for one-shot case. The foreground objects are combined with different backgrounds in attempt to refine the classification boundaries.}
    \label{fig:one-shot}
    \vspace{-0.4cm}
    \label{fig:illustration_method}
\end{figure}

In this paper, we adopt the data hallucination strategy and propose a saliency-guided data hallucination network dubbed as {\em Salient Network} ({\em SalNet}). Figure \ref{fig:illustration_method} shows a simple motivation for our work. Compared with previous feature hallucinating approaches, we employ a readily available saliency network \cite{Zhang_2018_CVPR} pre-trained on MSRA Salient Object Database (MSRA-B) \cite{Learning-Detect-Salient:CVPR-2007} to segment foregrounds and backgrounds from given images, followed by a two-stream network which mixes foregrounds with backgrounds (we call it the {\em Mixing Network}) in the feature space of an encoder (\cf image space). As we obtain spatial feature maps from this process, we embed mixed feature vectors into a second-order representation which aggregates over the spatial dimension of feature maps. Then, we capture the similarity between final co-occurrence descriptors of a so-called training query sample and hallucinated support matrices via a similarity-learning network. Moreover, we regularize our mixing network to promote hallucination of realistically blended foreground-background representations. To this end, whenever a foreground-background pair is extracted from the same image (\cf two separate images), we constrain the resulting blended representation via the $\ell_2$-norm to be close to a representation from a supervising network which, by its design, is trained only on real foreground-background pairs (\cf infeasible combinations). We refer to this strategy as {\em Real Representation Regularization} ({\em TriR}). Lastly, we propose the similarity-based strategies regarding how to choose backgrounds for mixing with a given foreground. To this end, we perform either (i) intra-class mixing (foregrounds/backgrounds of the same class) or (ii) inter-class mixing (for any given foreground, we take its corresponding background, retrieve its nearest-neighbour backgrounds from various classes, and use the retrieval distance to express the likelihood how valid the mixed pair is). 
Below, we list our contributions:
\renewcommand{\labelenumi}{\Roman{enumi}.}
\vspace{-2mm}
\hspace{-1cm}
\begin{enumerate}[leftmargin=0.6cm]
    \item We propose a novel saliency-guided data hallucination network for few-shot learning.
    \item We investigate various hallucination strategies. We propose a simple but effective regularization and  two strategies to prevent substandard hallucinated samples.
    \item We investigate the effects of different saliency map generators on the few-shot learning performance.
\end{enumerate}

To the best of our knowledge, we are the first to employ saliency maps for datapoints hallucination for few-shot learning. Our experiments achieve the state of the art on two challenging publicly available few-shot learning datasets.
\section{Related Work}
\label{sec:related}

In what follows, we describe popular zero-, one- and few-shot learning algorithms followed by the saliency detection methods and a discussion on second-order statistics.

\subsection{Learning From Few Samples}
\label{sec:related_few_shot}

For deep learning algorithms, the ability of {\em``learning quickly from only a few examples is definitely the desired characteristic to emulate in any brain-like system''} \cite{book_nip}. Learning from scarce data poses a challenge to typical CNN-based classification systems \cite{ILSVRC15} which have to learn millions of parameters. Current trends in computer vision highlight the need for {\em ``an ability of a system to recognize and apply knowledge and skills learned in previous tasks to novel tasks or new domains, which share some commonality''}. This problem was introduced in 1901 under a notion of ``{\em transfer of particle}''~\cite{woodworth_particle} and is closely related to zero-shot learning \cite{larochelle2008zero,farhadi2009describing,akata2013label} which can be defined as an ability to generalize to unseen class categories from categories seen during training. For one- and few-shot learning, some ``{\em transfer of particle}'' is also a desired mechanism as generalizing from one or few datapoints to account for intra-class variability of thousands images is a formidable task.

\noindent{\textbf{One- and Few-shot Learning }} has been  studied widely  in computer vision in both shallow \cite{miller_one_example,Li9596,NIPS2004_2576,BartU05,fei2006one,lake_oneshot} and deep learning scenarios \cite{koch2015siamese,vinyals2016matching,snell2017prototypical,finn2017model,snell2017prototypical,sung2017learning}. 

Early works \cite{fei2006one,lake_oneshot} propose generative models with an iterative inference for transfer. %in order to take advantage of knowledge from previously learned tasks. 
In contrast, a recent Siamese Network \cite{koch2015siamese} uses a two-stream convolutional neural network  which %generates image descriptors and 
performs simple metric learning. %The learning protocol learns to distinguish whether two datapoints come from the same or different class. %Therefore, one can see such an approach as similarity learning which is related to metric learning. 
Matching Network \cite{vinyals2016matching} introduces the concept of support set and $N$-way $W\!$-shot learning protocols. It  captures the similarity between one query and several support images, and also implicitly performs metric learning. %thus casting the one-shot learning problem as set-to-set learning. Such a network work with unobserved classes without any modifications. 
Prototypical Networks \cite{snell2017prototypical} learn a model that computes distances between a datapoint and prototype representations of each class. %
Model-Agnostic Meta-Learning (MAML) \cite{finn2017model} is a meta-learning model which can be seen a form of transfer learning. %as otherwise fine-tuning on few datapoints per classes is challenging. 
%Their model is trained on a variety of different learning tasks. %This results in a good initial condition for the solver to generalize to other novel tasks. % with few training samples. 
%
Relation Net \cite{sung2017learning} %proposes a simple but
is similar to Matching Network \cite{vinyals2016matching}, but uses an additional network to learn similarity between images. 
Second-order Similarity Network (SoSN) \cite{zhang2018power} leverages second-order descriptors and power normalization which help infer rich relation statistics. SoSN descriptors are more effective than the first-order Relation Net \cite{sung2017learning}. %Additionally they investigated various aggregations over features as well as the influences of different power normalizing functions.

Hallucination-based approaches \cite{hariharan2017low} and \cite{wang2018low} use %an additional network  trained on large numbers of samples, whose 
descriptors manually assigned into 100 clusters to  generate plausible combinations of datapoints. Mixup network \cite{zhang2018mixup} applies a convex combination of pairs of datapoints and labels. In contrast, we  decompose images into foreground and background representations via saliency maps and we propose several strategies for mixing foreground-background pairs to hallucinate meaningful auxiliary training samples.

\noindent{\textbf{Zero-shot Learning}} can be implemented within few-shot learning frameworks  \cite{koch2015siamese,vinyals2016matching,snell2017prototypical,sung2017learning}. %Methods such as 
Attribute Label Embedding (ALE) \cite{akata2013label}, %Embarrassingly Simple Zero-Shot Learning (ESZSL) \cite{romera2015embarrassingly}, 
Zero-shot Kernel Learning (ZSKL) \cite{zhang2018zero} all use so-called compatibility mapping (linear/non-linear) and some form of regularization to associate feature vectors with attributes (class descriptors). Recent methods such as Feature Generating Networks \cite{xian2017feature} and Model Selection Network \cite{zhang2018model} hallucinate the training data for unseen classes via Generative Adversarial Networks (GAN).
 %In contrast, our approach considers few-shot learning and saliency-guided mixture of foreground-background maps which bypasses costly training of GANs and avoids issues with their convergence.

\begin{figure*}[ht]
    \vspace{-0.5cm}
	\centering
	\includegraphics[width=0.9\linewidth]{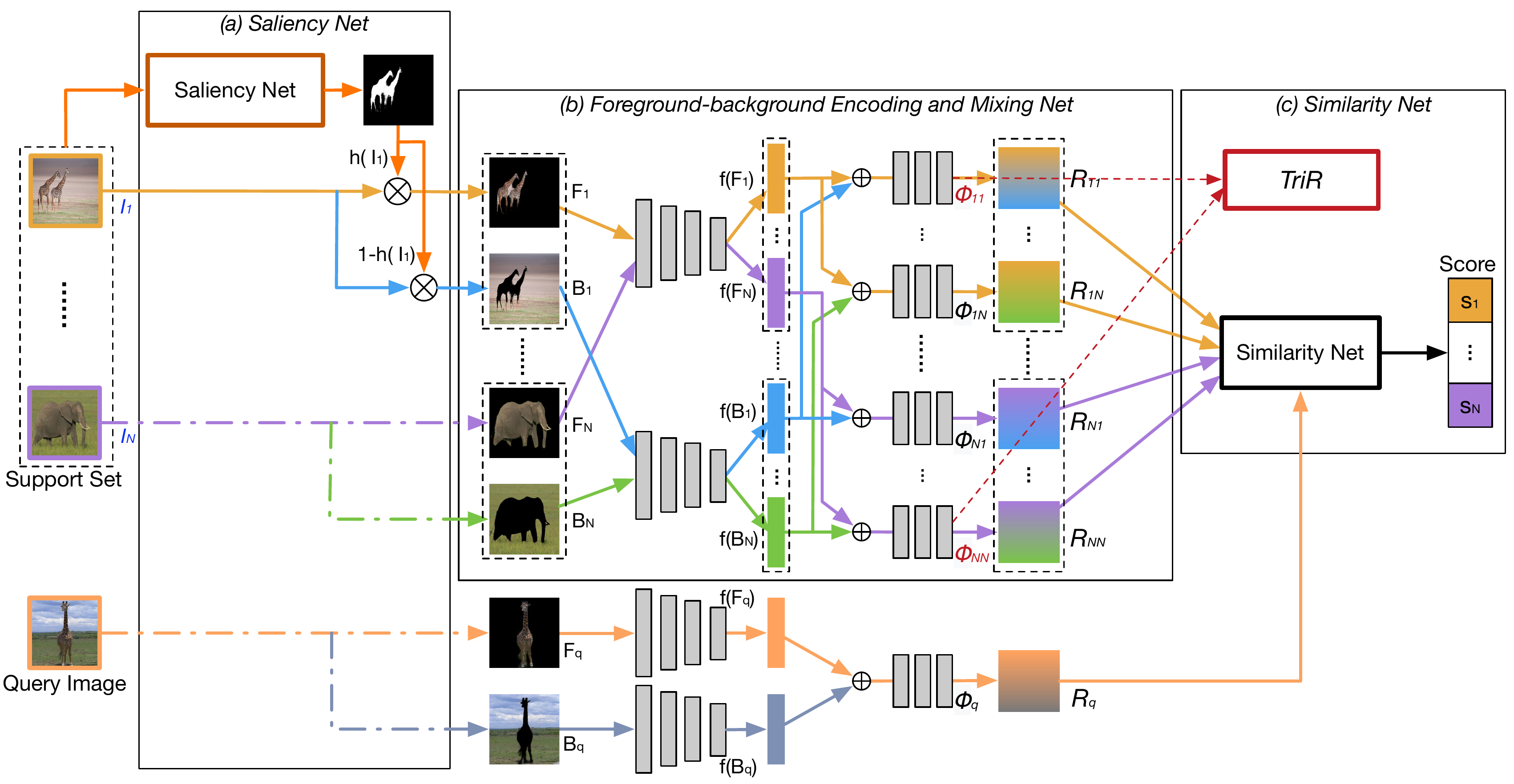}
	\caption{Our pipeline consists of three units: (a) pre-trained {\em Saliency Net}, (b) {\em Foreground-background Encoding and Mixing Net} ({\em FEMN}), and (c) {\em Similarity Net}. The FEMN block consists of two streams which take foreground/background images as inputs, respectively, and a {\em Mixing Net} which combines foreground-background pairs via $\oplus$ and refines them via a single-stream network prior to aggregation of the resulting feature maps via the {\em Second-order Encoder}. 
	%
	%\pk{Use {\em em} instead of bold to denote our components across the paper for the first time or in illsutrations. Give them capitalsied names \eg, {\em Saliency Net} ({\em SalNet}), {\em Similarity Net} ({\em SimNet}) } {\color{blue}{also, zoom TriR equation: there is room to the right of the red box!}}
	%Data hallucination incurs at the outputs of the first part of feature encoder only for support images. 
	}
	\label{fig:network}
	\vspace{-0.5cm}
\end{figure*}

%leverages attribute vectors as label embedding and uses an objective inspired by a structured WSABIE ranking method, which assigns more importance to the top of the ranking list. Embarrassingly Simple Zero-Shot Learning (ESZSL) \cite{romera2015embarrassingly} implements regularization terms for a linear mapping to penalize the projection of feature vectors to the attribute space and the projection of attribute vectors to the feature space. Zero-shot Kernel Learning (ZSKL) \cite{zhang2018zero} presents non-linear kernel methods which realize the compatibility function. The weak incoherence constraint is employed in this model to force the columns of projection matrix be incoherent, which acts as a regularization and makes their method related to subspace-learning approaches. 

%to hallucinate feature vectors for unseen classes, thus correcting the imbalance in the data distribution to improve results. %\cite{zhang2018model} proposes to train an extra SVM to detect if samples are from seen classes or unseen classes, thus employing different classifiers to classify images.
%\vspace{-0.2cm}
\subsection{Saliency Detection}
A saliency detector highlights image regions containing foreground objects which correlate with human visual attention, thus producing a dense likelihood saliency map which assigns  some relevance score in range $[0,1]$ to each pixel. Conventional saliency detectors underperform on complex scenes due to computations based on human-defined priors \cite{Background-Detection:CVPR-2014}. In contrast, deep saliency models \cite{RFCN,ChengCVPR17} outperform  conventional saliency detectors but they require laborious pixel-wise labels. In this paper, we use saliency maps as a guiding signal, thus we adopt a highly-efficient weakly-supervised deep convolutional saliency detector MNL \cite{Zhang_2018_CVPR}. We compare the performance of MNL with (i) RFCN \cite{RFCN}, a fully-supervised deep model, and (ii) a cheap non-CNN Robust Background Detector (RBD)  \cite{Background-Detection:CVPR-2014},  one of the best  unsupervised saliency detectors according to evaluation \cite{SalObjBenchmark_Tip2015}.

%Moreover, we use deep approaches such as RFCN \cite{RFCN} and MNL \cite{Zhang_2018_CVPR} for comparisons, where the former one is fully supervised deep model relying on pixel-wise labeling, and the later algorithm is does not use any human labeling.

\subsection{Second-order Statistics}
\label{sec:related_pn}

Below we discuss briefly the role of second-order statistics and related shallow and CNN-based approaches. % followed by details on so-called pooling and Power Normalization functions.

\noindent{\textbf{Second-order statistics}} have been studied in the context of texture recognition \cite{tuzel_rc, elbcm_brod} via so-called Region Covariance Descriptors (RCD), often  applied to semantic segmentation \cite{carreira_secord} and object category recognition \cite{me_tensor_tech_rep,koniusz2017higher}. %, to name but a few of applications.

%Recently, using co-occurrence patterns in CNN setting, similar in spirit to RCD, has become a popular direction. Approach \cite{bilinear_finegrained} fuses two CNN streams via outer product in the context of fine-grained image recognition. %Face recognition algorithm \cite{face_cooc} uses co-occurrences of CNN feature vectors and facial attribute vectors to obtain state-of-the-art face recognition results. 
%A recent approach \cite{deep_cooc} extracts feature vectors at two separate locations in feature maps and performs an outer product to form a CNN co-occurrence layer. 

Second-order statistics have to deal with the so-called burstiness which is ``{\em the property that a given visual element appears more times in an image than a statistically independent model would predict}'' \cite{jegou_bursts}. Power Normalization~\cite{me_ATN,me_tensor_tech_rep},   %, it is related to the problem of robust estimation of statistics, 
 used with Bag-of-Words  \cite{me_ATN,me_tensor_tech_rep,koniusz2017higher,koniusz2018deeper}, was shown to limit such a burstiness. 
%
%An analysis of pooling was conducted in~\cite{boureau_pooling} under  assumptions on distributions from which %the aggregated features are drawn.  %A relationship between the likelihood of `\emph{at least one particular visual word being present in an image}' and Max-pooling was studied in \cite{liu_sadefense}. 
%Later, 
A survey \cite{me_ATN} showed that so-called MaxExp feat. pooling \cite{boureau_pooling} is in fact a detector of ``\emph{at least one particular visual word being present in an image}''. %According to the survey \cite{me_ATN}, many Power Normalization functions are closely related. This view was further extended to
MaxExp on second-order matrices was shown in \cite{koniusz2018deeper} to be in fact the Sigmoid function. Such a pooling also performed well in few-shot learning \cite{zhang2018power}. Thus, we employ second-order pooling with Sigmoid.
\section{Approach}
Our pipeline builds on the generic few-shot Relation Net pipeline \cite{sung2017learning} which learns implicitly a metric for so-called query and support images. To this end, images are encoded into feature vectors by an encoding network. Then, so-called episodes with query and support images are formed. Each query-support pair is forwarded to a so-called relation network and a loss function to learn if a query-support pair is of the same class (1) or not (0). However, such methods suffer from scarce training data which we address below.

\subsection{Network}
Figure \ref{fig:network} presents a foreground-background two-stream network which leverages saliency maps to isolate foreground and background image representations %, thus improving the similarity comparing between images. 
in order to hallucinate additional training data to improve the few-shot learning performance. 
The network consists of (i) {\em Saliency Net} ({\em SalNet}) whose role is to generate foreground hypotheses, (ii) {\em Foreground-background Encoding and Mixing Net} ({\em FEMN}) whose role is to combine foreground-background image pairs into episodes, and the {\em Similarity Net} ({\em SimNet}) which learns the similarity between query-support pairs.

To illustrate how our network works, consider an image $\mathbf{I}$ which is passed through some saliency network $\mathbf{h}$ to extract the corresponding saliency map $\mathbf{h}(\mathbf{I})$, %$\mathbf{S}$ firstly. Then we use the saliency map to obtain 
the foreground $\mathbf{F}$ and the background $\mathbf{B}$ of $\mathbf{I}$, respectively:

\vspace{-0.5cm}
\begin{align}
&\mathbf{F}_{I} = \mathbf{h}(\mathbf{I})\odot \mathbf{I},\\
&\mathbf{B}_{I} = (1 - \mathbf{h}(\mathbf{I}))\odot\mathbf{I},
\end{align}
%\vspace{-0.2cm}    
%
where $\odot$ is the Hadamart product. The feature encoding network consists of two parts, $\mathbf{f}$ and $\mathbf{g}$.
For images $\mathbf{I}\!\in\!\mbr{3\times M\times M}$ and $\mathbf{J}\!\in\!\mbr{3\times M\times M}$ ($\mathbf{I}\!=\!\mathbf{J}$ or $\mathbf{I}\!\neq\!\mathbf{J}$), we proceed by encoding their foreground $\mathbf{F}_{I}\!\in\!\mbr{3\times M\times M}$ and background $\mathbf{B}_{J}\!\in\!\mbr{3\times M\times M}$ via feature encoder $\mathbf{f}: \mbr{3\times M\times M}\!\!\!\rightarrow\!\mbr{K\times Z^2}$, where $M\!\times\!M$ denotes the spatial size of an image, $K$ is the feature size and $Z^2$ refers to the vectorized spatial dimension of map of $\mathbf{f}$ of size $Z\!\times\!Z$. Then, the encoded foreground and background are mixed via summation and refined in encoder $\mathbf{g}: \mbr{K\times Z^2}\!\!\!\rightarrow\!\mbr{K'\times Z'^2}$, where $K'$ is the feature size and $Z'^2$ corresponds to the vectorized spatial dimension of map of $\mathbf{g}$ of size $Z'\!\!\times\! Z'$. As in the SoSN approach \cite{zhang2018power}, we apply the outer-product on $\mathbf{g}(\cdot)$ to obtain an auto-correlation of features and we perform pooling via Sigmoid $\vpsi$ to tackle the burstiness in our representation. Thus, we have:
\vspace{-0.1cm}
\begin{align}
& \mPhi_{I\!J}=\mathbf{g}(\mathbf{f}(\mathbf{F}_{I}) + \mathbf{f}(\mathbf{B}_{J})),\\
&\mathbf{R}_{I\!J} = \vpsi (\mPhi_{I\!J}\mPhi_{I\!J}^T,\, \sigma),
\vspace{-0.1cm}
\end{align}
where $\vpsi$ is a zero-centered Sigmoid function  with $\sigma$ as the parameter that controls the slope of its curve:
\vspace{-0.1cm}
\begin{equation}
    \vpsi(\mX,\sigma) = %\frac{1 - e^{-\sigma \mX}}{1+e^{-\sigma \mX}}.
    ({1\!-\!e^{-\sigma \mX}})/({1\!+\!e^{-\sigma \mX}})=\text{tanh}(2\sigma\mX).
\vspace{-0.1cm}
\end{equation}
Descriptors $\mathbf{R}_{I\!I}\!\in\!\mbr{K'\times K'}$ represent a given image $\mathbf{I}$ while $\mathbf{R}_{I\!J}\!\in\!\mbr{K'\times K'}$ represent a combined foreground-background pair of images $\mathbf{I}$ and $\mathbf{J}$.
Subsequently, we form the query-support pairs (\eg,  we concatenate their representations) and we pass episodes to the similarity network. % to learn the relation between images.
We use the Mean Square Error (MSE) loss to train our network:
\vspace{-0.2cm}
\begin{equation}
    L = \frac{1}{NW} \sum\limits_{n=1}^N \sum\limits_{w=1}^W(r(\mathbf{R}_{s_{nw}}, \mathbf{R}_q) -\delta(l_{s_{nw}}\!-l_q))^2,
    \label{eq:loss1}
    %\vspace{-0.1cm}
\end{equation}
where $s_{nw}$ chooses support images from $\mathcal{I}\!=\!\mathcal{I}^*\!\!+\!\mathcal{I}'\!$, $\mathcal{I}^*\!$ and $\mathcal{I}'\!$ are original and hallucinated images, $q$ chooses the query image, $r$ is the similarity network, $l$ is the label of an image, $N$ is the number of classes in an episode, $W$ is the shot number per support class, $\delta(0)\!\!=\!\!1$ (0 elsewhere). Note that Eq. \eqref{eq:loss1} does not form  foreground-background hallucinated pairs per se. We describe this process in Section \ref{sec:hall}. % and provide our updated loss.

\subsection{Saliency Map Generation}
For brevity, we consider three approaches: deep supervised saliency approaches \cite{Zhang_2018_CVPR, RFCN} and an unsupervised shallow method \cite{Background-Detection:CVPR-2014}. In this paper, we use saliency maps as a prior to generate foreground and background hypotheses.

In our main experiemnts, we use the deep weakly-supervised slaiency detector MNL \cite{Zhang_2018_CVPR} due to its superior performance. Moreover, we investigate the deep supervised RFCN approach \cite{RFCN} pre-trained on THUS10K dataset \cite{Global-Contrast:CVPR-2011}, which has no intersection with our few-shot learning datasets. We also investigate the cheap RBD model \cite{Background-Detection:CVPR-2014} which performed best among unsupervised models \cite{SalObjBenchmark_Tip2015}.

\begin{figure}[t]
\hspace{-0.2cm}
\vspace{-0.1cm}
   %\begin{center}
   \begin{tabular}{ c@{ } c@{ } c@{ } c@{ }}
   {\includegraphics[width=0.24\linewidth]{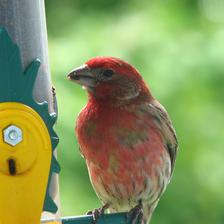}}&
   {\includegraphics[width=0.24\linewidth]{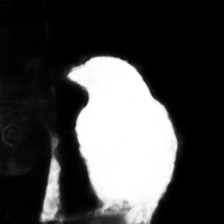}}&
   {\includegraphics[width=0.24\linewidth]{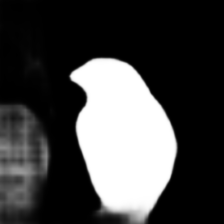}}&
   {\includegraphics[width=0.24\linewidth]{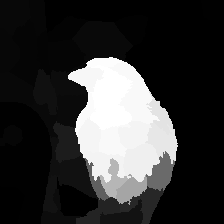}} \\
   {\includegraphics[width=0.24\linewidth]{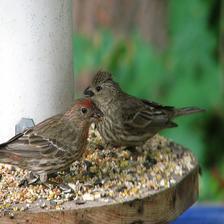}}&
   {\includegraphics[width=0.24\linewidth]{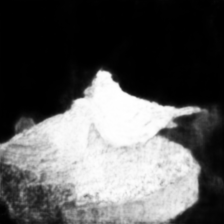}}&
   {\includegraphics[width=0.24\linewidth]{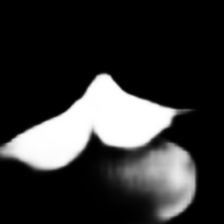}}&
   {\includegraphics[width=0.24\linewidth]{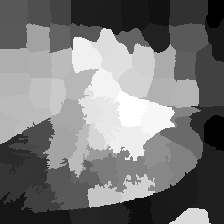}} \\
   \footnotesize{\textbf{Image}} & \footnotesize{\textbf{RFCN}} \cite{RFCN} & \footnotesize{\textbf{MNL}} \cite{Zhang_2018_CVPR} & \footnotesize{\textbf{RBD}} \cite{Background-Detection:CVPR-2014} \\
   \end{tabular}
   %\end{center}
   \vspace{-3mm}
   \caption{Saliency maps generated by different methods. For a simple scene (top row), the all three methods are able to detect the foreground. However, for a complex scene, the unsupervised method fails to detect the salient object.}
   \label{fig:sal_map}
   \vspace{-0.4cm}
\end{figure}

Figure \ref{fig:sal_map} shows saliency maps generated by the above methods. In the top row, the foreground and background have distinct textures. Thus, both conventional and deep models isolate the foreground well. However, for the scenes whose foreground/background share color and texture composition (bottom row), the unsupervised method fails to detect the correct foreground. As our dataset contains both simple and complex scenes, the performance of our method is somewhat dependent on the saliency detector \eg, results based on RBD \cite{Background-Detection:CVPR-2014} are expected to be worse in comparison to RFCN \cite{RFCN} and MNL \cite{Zhang_2018_CVPR}. The performance of few-shot learning combined with different saliency detectors will be presented in Section \ref{sec:resres}. Firstly, we detail our strategies for hallucinating additional training data for few-shot learning.

\begin{figure*}[t]
\vspace{-0.6cm}
	\centering
	\includegraphics[width=0.9\linewidth]{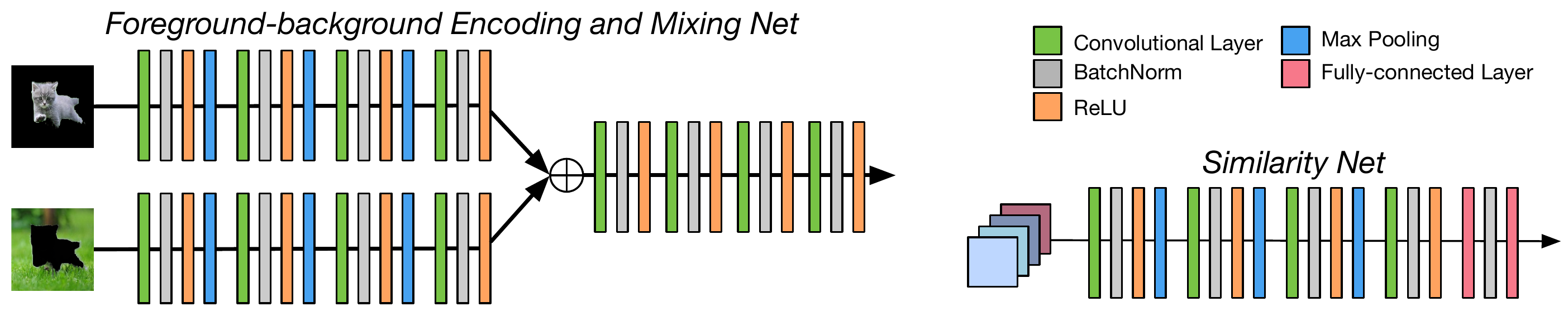}
	\caption{The detailed architecture of Foreground-Background Encoding and Mixing Net and the Similarity Net. Best viewed in color.}
	\label{fig:blocks}
	\vspace{-0.4cm}
	\label{fig:detail_net_view}
\end{figure*}

\subsection{Data Hallucination} 
\label{sec:hall}

The additional datapoints are hallucinated by the summation of foreground and background feature vector pairs obtained from the feature encoder $\mathbf{f}$ and refined by the encoder $\mathbf{g}$. 
Taking the $N$-way $W\!$-shot problem as example (see Relation Net \cite{sung2017learning} or SoSN \cite{zhang2018power} for the detailed definition of such a protocol), we will randomly sample $W$ images from each of $N$ training classes. Let $s_{nw}$ be the index selecting the $w$-th image from the $n$-th class of an episode and $q$ be the index selecting the query image. Where required, assume the foreground and background descriptors for images are extracted. Then, the following strategies for  the hallucination of auxiliary datapoints can be formulated.

\begin{figure}[h]
%\vspace{-0.3cm}
	\centering
	\includegraphics[width=7.5cm]{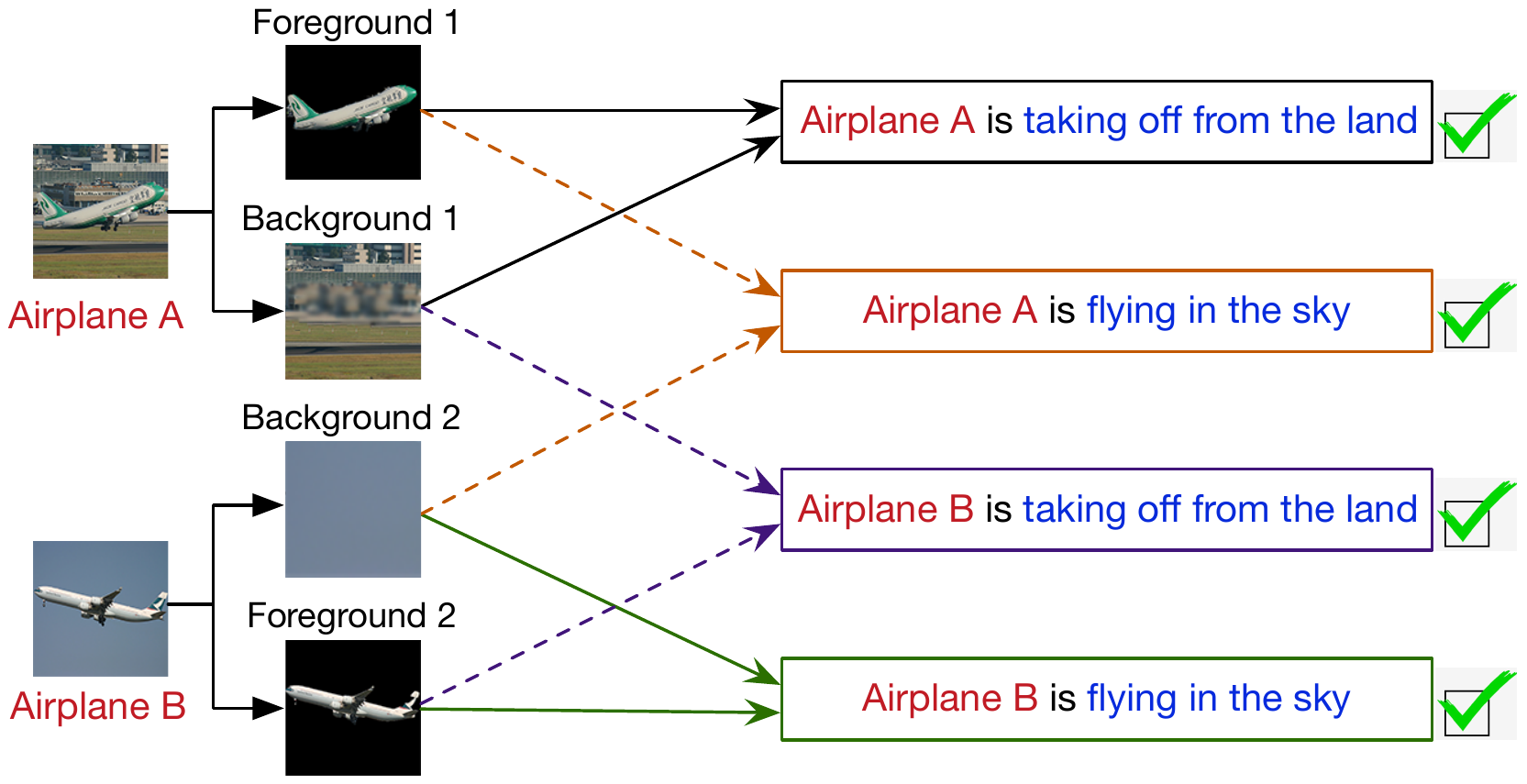}
	\caption{\small The intra-class datapoint hallucination strategy: the majority of datapoints generated in this way are statistically plausible.}
	\label{fig:intraclass}
	\vspace{-0.3cm}
\end{figure}

\vspace{0.1cm}
\noindent{\textbf{Strategy \uppercase\expandafter{\romannumeral1}: Intra-class hallucination.}} For this strategy, given an image index $s_{nw}$, a corresponding foreground is only mixed with backgrounds of images from the same class $n$. Thus, we can generate $W\!-\!1$ datapoints for every image. Figure \ref{fig:intraclass} shows that {\em the intra-class hallucination} produces plausible new datapoints. Note that the image class $n$ typically correlates with foreground objects, and such objects appear on backgrounds which, statistically speaking, if swapped, will  produce plausible object-background combinations. However,
the above strategy cannot work in one-shot setting as only one support image per class is given.

Although our intra-class hallucination presents a promising direction, our results will show that sometimes the performance may lie below  baseline few-shot learning due to a very simple mixing foreground-background strategy which includes the foreground-background feature vector summation followed by the refining encoder $\mathbf{g}$. %some implausible foreground-background combinations.
Such a strategy incurs possible noises from (i) the substandard saliency maps and/or (ii) mixing incopatible foreground-background pairs.

%It seems that the feature encoding network trained with hallucination generates incorrect representations in the end-to-end training manner. 
Therefore, in order to further refine the hallucinated datapoints, we propose to exploit foreground-background mixed pairs $\mathbf{F}_{s_{nw}}$ and $\mathbf{B}_{s_{nw}}$ which come from the same image (\eg, their mixing should produce the original image) and enforce their feature vectors to be close in the $\ell_2$-norm sense to some baseline teacher network which does not perform hallucination. Specifically, we take $\mPhi\!=\!\mathbf{g}(\mathbf{F}_{s_{nw}}, \mathbf{B}_{s_{nw}})$ and encourage its proximity to some teacher representation $\mPhi^*\!\!=\!\mathbf{g}^*(\{\mathbf{F}_{s_{nw}}, \mathbf{B}_{s_{nw}}\})$ where $\mathbf{F}_{s_{nw}}\!\!+\!\mathbf{B}_{s_{nw}}\!\!=\!\mathbf{I}_{s_{nw}}\!\!\in\!\mathcal{I}^*$:

%we pass the aggregated intermediate representations of {\em 'no-hallucination'} combinations, which denote the original pairs of foreground $\mathbf{F}_{s_{ij}}$ and background $\mathbf{B}_{s_{ij}}$, to encoding network $g^*$, 
\vspace{-0.3cm}
\begin{align}
& \Omega\!=\!\Scale[0.8]{\frac{1}{N W}} \sum\limits_{n=1}^{N} \sum\limits_{w=1}^{W}\Scale[0.8]{\Big\lVert\mathbf{g}(\mathbf{f}(\mathbf{F}_{s_{nw}})+\mathbf{f}(\mathbf{B}_{s_{nw}})) - \mathbf{g}^*(\{\mathbf{F}_{s_{nw}}, \mathbf{B}_{s_{nw}} \})\Big\rVert_2^2},\nonumber\\
& \qquad\text{s.t. } \mathbf{F}_{s_{nw}}\!\!+\!\mathbf{B}_{s_{nw}}\!\!=\!\mathbf{I}_{s_{nw}}\!\!\in\!\mathcal{I}^*\label{eq:trir}\\
&L' = L + \beta \Omega,\nonumber
\end{align}
where $\mathcal{I}^*\!$ is a set of orig. train. images, $\beta$ adjusts the impact of $\Omega$, $L'\!$ is the combined loss, and net. $\mathbf{g}^*\!$ is already trained.

We investigate $\mathbf{g}^*\!$ that encodes (i) the original images only \ie, $\mathbf{g}^*\!(\mathbf{f}(\mathbf{I}_{nw}))$ or (ii) foreground-background pairs from original images \ie, $\mathbf{g}^*\!(\mathbf{f}(\mathbf{F}_{s_{nw}})\!+\!\mathbf{f}(\mathbf{B}_{s_{nw}})$. We call $\Omega$ as {\em Real Representation Regularization} ({\em TriR}). Our experiments will demonstrate that TriR improves the final results. %hallucinated representations.

\begin{figure}[t]
	%\vspace{-0.5cm}
	\centering
	\includegraphics[width=7.5cm]{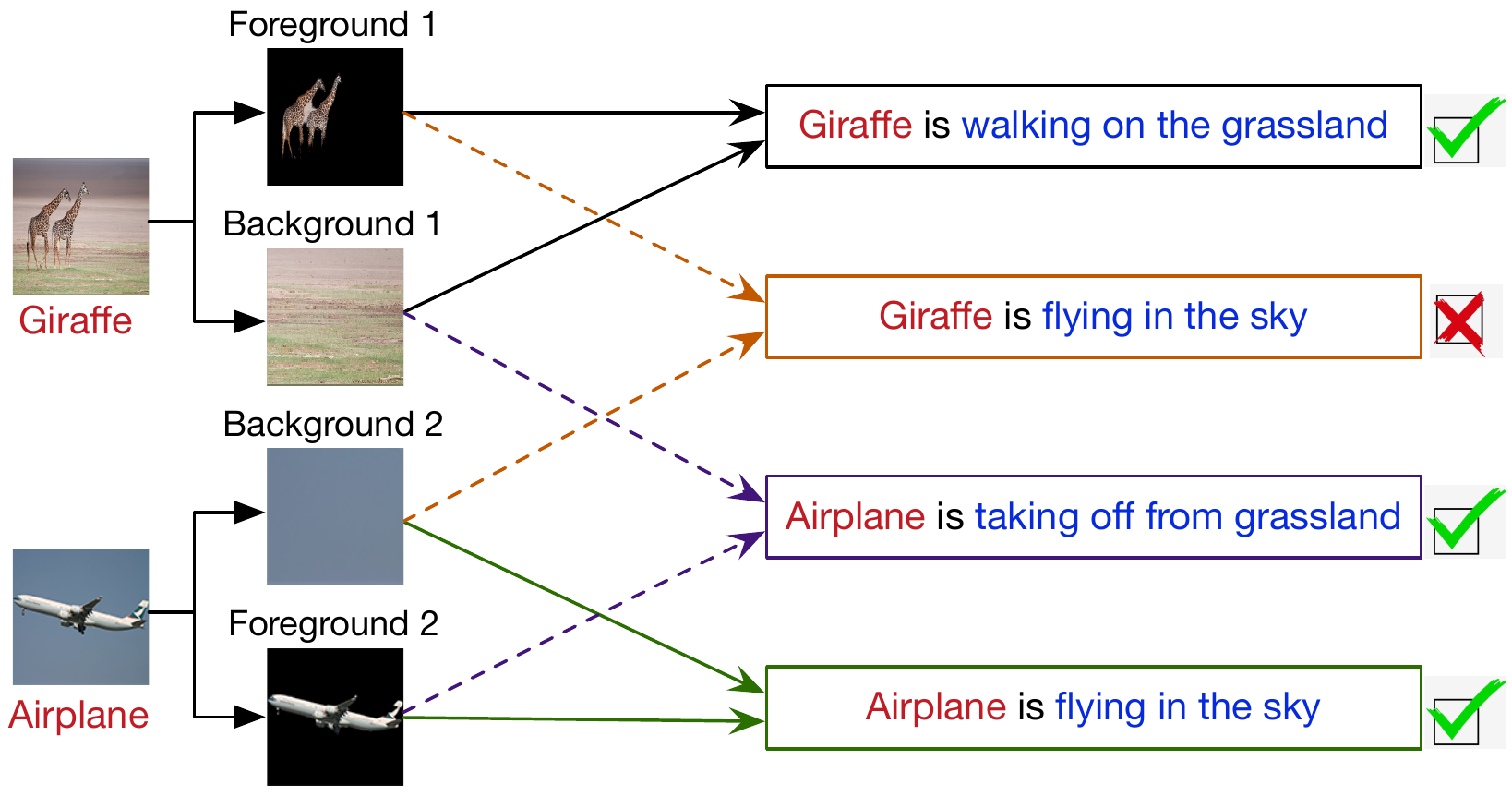}
	\caption{\small The inter-class datapoint hallucination may generate impossible instances \eg, `giraffe in the sky' is an unlikely concept (except for a giraffe falling off a helicopter during transportation?).}
	\label{fig:interclass}
	\vspace{-0.3cm}
\end{figure}

\vspace{0.1cm}
\noindent{\textbf{Strategy \uppercase\expandafter{\romannumeral2}: Inter-class hallucination.}} For this strategy, we allow mixing the foregrounds of support images with all available backgrounds (between-class mixing is allowed) in the support set. Compared to the intra-class generator, {\em the inter-class hallucination} can generate $W\!-\!1\!+\!W(N\!-\!1)$ new datapoints. However, many  foreground-background pairs will be statistically implausible, as shown in Figure \ref{fig:interclass}, which would cause the degradation of the classification accuracy.

To eliminate the implausible foreground-background pairs from the inter-class hallucination process, we design a similarity prior which assigns probabilities to backgrounds in terms of their compatibility with a given foreground.%to alleviate the noises in inter-class scenario.

\begin{figure}[t]
\vspace{-0.3cm}
	\centering
	\includegraphics[width=0.8\linewidth]{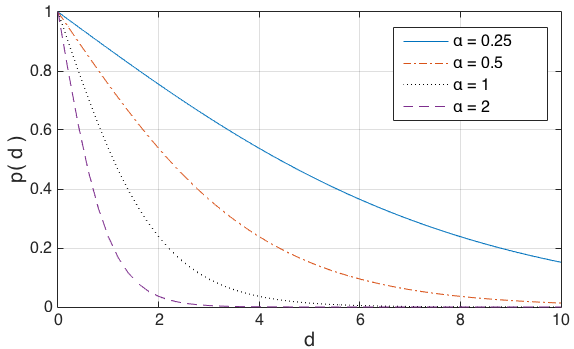}
	\vspace{-0.3cm}
	\caption{\small The probability profile $p$ w.r.t. the dist. $d$ and various  $\alpha$.}
	%\vspace{-0.1cm}
	\label{fig:probability_profile}
\end{figure}

\begin{figure}[t]
\vspace{-0.1cm}
	\centering
	\includegraphics[width=7.5cm]{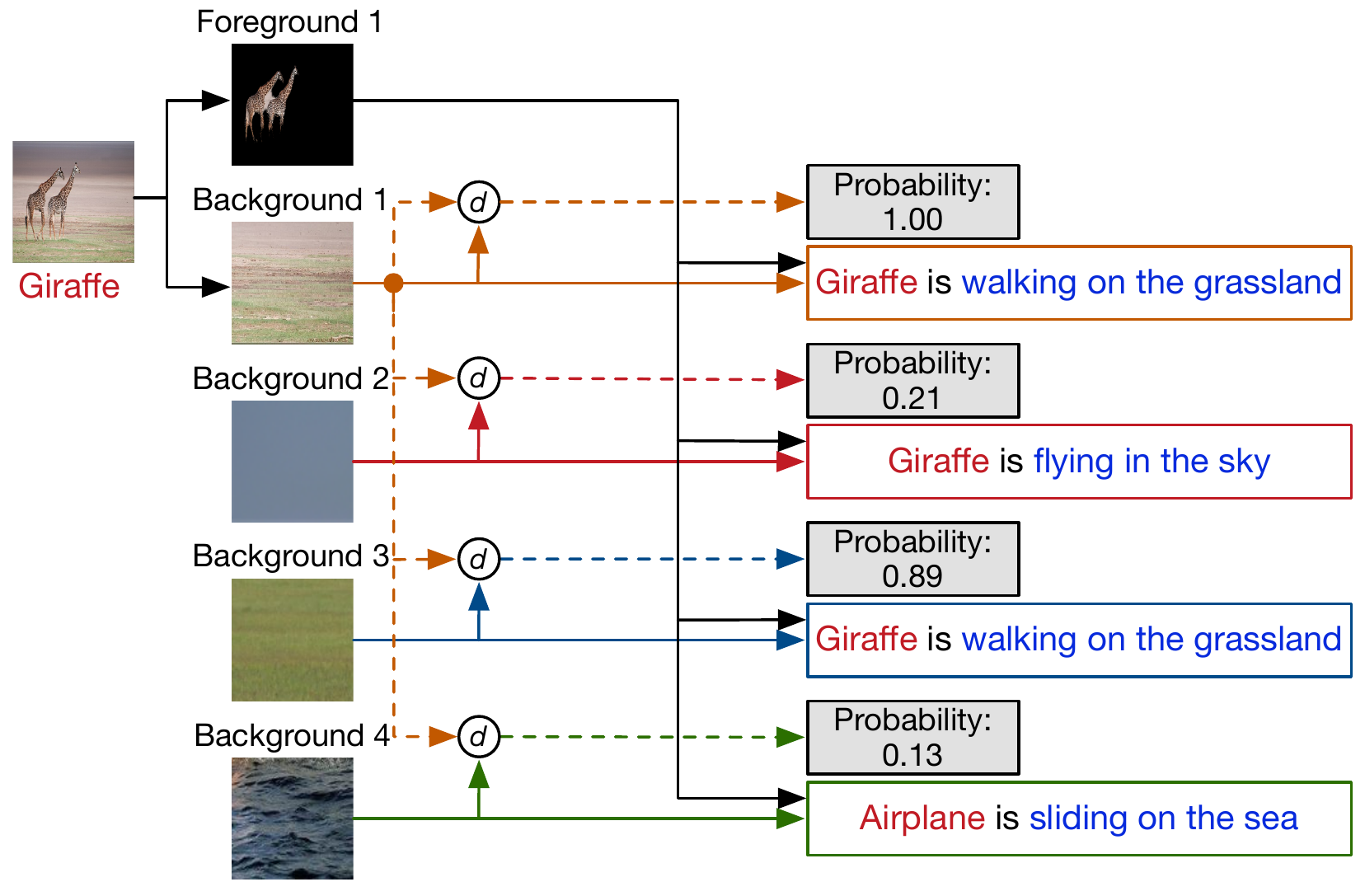}
	\caption{\small The inter-class hallucination strategy with the similarity prior. We assign likelihoods to generated datapoints based on the similarity of a background of a given image to other backgrounds.}
	\label{fig:simprior}
	\vspace{-0.4cm}
\end{figure}

Numerous similarity priors can be proposed \eg, one can use the label information to specify some similarity between two given classes. Intuitively, backgrounds between images containing dogs and cats should be more correlated than backgrounds of images of dogs and radios. However, modeling such relations explicitly may be cumbersome and it has its shortcomings \eg, backgrounds of images containing cars may also be suitable for rendering animals on the road or sidewalk, despite of an apparent lack of correlation between say cat and car classes. Thus, we ignore  class labels and perform a background retrieval instead. Specifically, once all backgrounds of support images are extracted, we measure the distance between the background of a chosen image of index $s_{nw}$  and all other backgrounds to assign a probability score of how similar two backgrounds are, thus:
\begin{align}
\vspace{-0.1cm} 
&d(\mathbf{B}_{s_{nw}}, \mathbf{B}_{s_{n'w'\!}}) = \big\lVert \mathbf{f}(\mathbf{B}_{s_{nw}}) - \mathbf{f}(\mathbf{B}_{s_{n'w'\!}})\big\rVert_2^2,\\
&p(\mathbf{B}_{s_{n'w'\!}}|\,\mathbf{B}_{s_{nw}}) = \frac{2e^{-\alpha d(\mathbf{B}_{s_{nw}}, \,\mathbf{B}_{s_{n'w'\!}})}}{1 + e^{-\alpha d(\mathbf{B}_{s_{nw}}, \,\mathbf{B}_{s_{n'w'\!}})}},
	\vspace{-0.1cm}     
\end{align}
where $\alpha$ is a hyper-parameter to control our probability profile function $p(d)$ shown in Figure \ref{fig:probability_profile}: a Sigmoid reflected along its y axis. 
We apply the profile $p$ to hallucinated outputs of $\mathbf{g}$ to obtain $\mathbf{g}'\!$. We show this strategy in Figure \ref{fig:simprior} and we call it as {\em Soft Similarity Prior} ({\em SSP}):
\begin{equation}
	%\vspace{-0.2cm} 
    \mathbf{g}'(\mathbf{F}_{s_{nw}}, \mathbf{B}_{s_{n'w'\!}})\!=\! p(\mathbf{B}_{s_{n'w'\!}}|\,\mathbf{B}_{s_{nw}})\,\mathbf{g}(\mathbf{f}(\mathbf{F}_{s_{nw}}), \mathbf{f}(\mathbf{B}_{s_{n'w'\!}})).
    \label{eq:ssp}
	%\vspace{-0.2cm}    
\end{equation}

Also, we propose a {\em Hard Similarity Prior} ({\em HSP}) according to which we combine a given foreground with the most relevant retrieved backgrounds whose $p$ is above certain $\tau$:

\vspace{-0.2cm}
\begin{equation}
%\mathbf{g}'(\mathbf{f}(\mathbf{F}_{s_{nw}}, \mathbf{B}_{s_{n'w'\!}}))=\left\{
\mathbf{g}'(\mathbf{F}_{s_{nw}}, \mathbf{B}_{s_{n'w'\!}})=\left\{
\begin{aligned}
\mathbb{0}, \quad \textbf{if} \,\, p(\mathbf{B}_{s_{n'w'\!}}|\mathbf{B}_{s_{nw}}) \leq \tau, \\
\mathbf{g}(\mathbf{f}(\mathbf{F}_{s_{nw}}), \mathbf{f}(\mathbf{B}_{s_{n'w'\!}})),  \quad \text{otherwise.}
\end{aligned}
\right.
\end{equation}

We will show in our experiments that the use of priors significantly enhances the performance of the inter-class hallucination, especially for the 1-shot protocol, to which the intra-class hallucination  is not applicable. We will show in Section \ref{sec:exp} that both HSP and SSP improve the performance of few-shot learning; SSP being a consistent performer on all protocols. Firstly, we detail datasets  and then we show the usefulness of our  approach experimentally.

\section{Experiments}
\label{sec:exp}
 Our network is evaluated in the few-shot learning scenario on the \textit{mini}Imagenet \cite{vinyals2016matching} dataset and a recently proposed Open MIC dataset \cite{me_museum} which was used for few-shot learning by the SoSN approach \cite{zhang2018power}. Our implementation is based on PyTorch and models are trained on a Titan Xp GPU via %Nvidia Tesla P100 GPU.  
the Adam solver. The architecture of our saliency-guided hallucination network is shown in Fig. \ref{fig:network} and \ref{fig:blocks}. The results are compared with several state-of-the-art methods for one- and few-shot learning.

\subsection{Datasets}
Below, we describe our  setup, datasets and evaluations.

%\vspace{0.05cm}
\noindent\textbf{\textit{mini}Imagenet} \cite{vinyals2016matching} consists of 60000 RGB images from 100 classes. %, each class containing 600 samples. 
We follow the standard protocol \cite{vinyals2016matching} and use 80 classes for training (including 16 classes for validation) and  20 classes for testing. All images are resized to $84\!\times\!84$ pixels for fair comparison with other methods. We also investigate larger sizes, \eg~$224\!\times\!224$, as our SalNet model can use richer spatial information from larger images to obtain high-rank auto-correlation matrices without a need to modify the similarity network to larger feature maps.

%In contrast, increasing image size in  Relation Net \cite{sung2017learning} results in a large number of extra convolutional blocks need to to be added to the image descriptor and/or similarity network to deal with the increase in the spatial dimension.

\begin{table}[t]
\vspace{-0.3cm}
\centering
\caption{\small Evaluations on the \textit{mini}Imagenet dataset. See \cite{sung2017learning, zhang2018power} for details of baselines. Note that intra-class hallucination has no effect on one-shot learning, so the scores of without  ({\em w/o Hal.}) and with intra-class hallucination ({\em Intra-class Hal.}) on 1-shot are the same. The astersik ({\em *}) denotes the `sanity check' results on our proposed pipeline given disabled both saliency segmentation and hallucination (see the supp. material for details).}
\label{table2}
\vspace{-0.1cm}
%\makebox[\linewidthwidth]{
\setlength{\tabcolsep}{0.10em}
\renewcommand{\arraystretch}{0.70}
%\fontsize{8.5}{9}\selectfont
\begin{tabular}{l|c|c|c} 
 & Fine & \multicolumn{2}{c}{5-way Acc.} \\ 
Model & Tune & 1-shot & 5-shot \\ \hline
\textit{Matching Nets} \cite{vinyals2016matching}& N & \,\,$43.56 \pm 0.84$\,\, & \,\,$55.31 \pm 0.73$\,\,  \\
\textit{Meta Nets} \cite{munkhdalai2017meta} & N & $49.21 \pm 0.96$ & - \\
\textit{Meta-Learn Nets} \cite{ravi2016optimization} & N & $43.44 \pm 0.77$ & $60.60 \pm 0.71$ \\
\textit{Prototypical Net} \cite{snell2017prototypical}& N & $49.42 \pm 0.78$ & $68.20 \pm 0.66$ \\ 
\textit{MAML} \cite{finn2017model}& Y & $48.70 \pm 1.84$ & $63.11 \pm 0.92$ \\ 
\textit{Relation Net}  \cite{sung2017learning}& N & $51.36 \pm 0.86$ & $65.63 \pm 0.72$  \\
\textit{SoSN} \cite{zhang2018power}& N & $52.96 \pm 0.83$ & $68.63 \pm 0.68$ \\
\hline
\textit{SalNet {\small w/o Sal. Seg.}} (*) & N & $ 53.15\pm 0.87$ & $ 68.87 \pm 0.67$ \\
\textit{SalNet {\small w/o Hal.}} & N & \multirow{2}{*}{$55.57 \pm 0.86$} & $70.35 \pm 0.66$ \\
\textit{SalNet {\small Intra. Hal.}} & N &  & $ 71.78\pm 0.69$ \\
%\textit{SalNet {\small Intra. Hal. Dilation}} & N & $ 56.85 \pm 0.87$ & $ 71.83\pm 0.67$ \\
\textit{SalNet {\small Inter. Hal.}} & N & $ 57.45 \pm 0.88$ & $ 72.01\pm 0.67$ \\
%\textit{SalNet {\small Inter. Hal. Dilation}} & N & $ 57.71 \pm 0.89$ & $ 72.12\pm 0.68$ \\
\hline
\end{tabular}
\vspace{-0.2cm}
\end{table}

\noindent\textbf{Open MIC} is a recently proposed Open Museum Identification Challenge (Open MIC) dataset \cite{me_museum} which contains photos of various exhibits, \eg~paintings, timepieces, sculptures, glassware, relics, science exhibits, natural history pieces, ceramics, pottery, tools and indigenous crafts, captured from 10 museum exhibition spaces according to which it is divided into 10 subproblems. In total, Open MIC has 866 diverse classes and 1--20 images per class. The within-class images undergo various geometric and photometric distortions as the data was captured with wearable cameras. This makes Open MIC a perfect candidate for testing one-shot learning algorithms. Following the setup in SoSN \cite{zhang2018power}, we combine ({\em shn+hon+clv}), ({\em clk+gls+scl}), ({\em sci+nat}) and ({\em shx+rlc}) splits into subproblems {\em p1}, $\!\cdots$, {\em p4}. We randomly select 4 out of 12 possible pairs in which subproblem $x$ is used for training and $y$ for testing (x$\rightarrow$y).

Relation Net \cite{sung2017learning} and SoSN \cite{zhang2018power} are employed as baselines against which we compare our SalNet approach. 

\subsection{Experimental setup}
%\vspace{-0.2cm}
For the {\em mini}Imagenet dataset, we perform 1- to 10-shot experiments in 5-way scenario to demonstrate the improvements obtained with our SalNet on different number of $W\!$-shot images. For every training and testing episode, we randomly select 5 and 3 query samples per class. We average the final results over 600 episodes. The initial learning rate is set to $1e\!-\!3$. We train the model with $200000$ episodes.

For the Open MIC dataset, we select 4 out of 12 possible subproblems, that is $p1\!\!\rightarrow\!p2$, $p2\!\!\rightarrow\!p3$, $p3\!\!\rightarrow\!p4$, and $p4\!\!\rightarrow\!p1$. Firstly, we apply the mean extraction on patch images (Open MIC provides three large crops per image) and resize them to $84\!\times\!84$ pixels. As some classes of Open MIC contain less than 3 images, we apply 5-way 1-shot to 3-shot learning protocol. During training, to form an episode, we randomly select 1--3 patch images for the support set and another 2 patch images for the query set for each class. During testing, we use the same number of support and query samples in every episode and we average the accuracy over 1000 episodes for the final score. The initial learning rate is set to $1e\!-\!4$. The models are trained with $50000$ episodes.
%------------------------------------------------------

\begin{table}[t]
\vspace{-0.3cm}
%\centering
\caption{\small Evaluations on the Open MIC dataset. p1: shn+hon+clv, p2: clk+gls+scl, p3: sci+nat, p4: shx+rlc. Notation $x\!\!\rightarrow\!y$ means training on exhibition $x$ and testing on exhibition $y$.}
\label{table5}
%\vspace{-0.1cm}
%\makebox[\linewidth]{
\setlength{\tabcolsep}{0.20em}
\renewcommand{\arraystretch}{0.70}
\fontsize{9}{10}\selectfont
\hspace{-0.3cm}
\begin{tabular}{l|c|c|c|c|c|c}
Model & $N$-way & $W\!$-shot & $p1\!\!\rightarrow\!p2$ & $p2\!\!\rightarrow\!p3$ & $p3\!\!\rightarrow\!p4$ & $p4\!\!\rightarrow\!p1$\\ \hline
\small Relation Net\cite{sung2017learning} & 5 & 1 & 70.1 & 49.7 & 66.9 & 46.9\\
\small SoSN \cite{zhang2018power}          & 5 & 1 & 78.0 & 60.1 & 75.5 & 57.8\\
\arrayrulecolor{black} \cdashline{1-7}[1pt/3pt]
\textit{Intra.-Hal.}                       & 5 & 1 & 78.2 & 60.3 & 75.9 & 58.1 \\
\textit{Inter.-Hal.}                       & 5 & 1 & 79.3 & 61.4 & 76.6 & 59.2\\
\hline
\small Relation Net\cite{sung2017learning} & 5 & 2 & 75.6 & 55.2 & 72.3 & 56.0 \\
\small SoSN \cite{zhang2018power}          & 5 & 2 & 84.6 & 68.1 & 82.7 & 66.8\\
\arrayrulecolor{black} \cdashline{1-7}[1pt/3pt]
\textit{Intra.-Hal.}                       & 5 & 2 & 85.7 & 69.2 & 84.1 & 67.5 \\
\textit{Inter.-Hal.}                       & 5 & 2 & 86.4 & 70.0 & 84.3 & 67.8\\
\hline
\small Relation Net\cite{sung2017learning} & 5 & 3 & 80.9 & 61.9 & 78.5 & 58.9 \\
\small SoSN \cite{zhang2018power}          & 5 & 3 & 87.1 & 72.6 & 85.9 & 72.8\\
\arrayrulecolor{black} \cdashline{1-7}[1pt/3pt]
\textit{Intra.-Hal.}                       & 5 & 3 & 87.5 & 73.9 & 86.5 & 73.6\\
\textit{Inter.-Hal.}                       & 5 & 3 & 88.1 & 74.2 & 87.1 & 73.9\\
\hline
\small Relation Net\cite{sung2017learning} & 10 & 1 & 54.4 & 35.3 & 53.1 & 35.5\\
\small SoSN \cite{zhang2018power}          & 10 & 1 & 67.2 & 46.2 & 63.9 & 46.6\\
\arrayrulecolor{black} \cdashline{1-7}[1pt/3pt]
\textit{Intra.-Hal.}                       & 10 & 1 & 67.6 & 46.7 & 64.3 & 47.0\\
\textit{Inter.-Hal.}                       & 10 & 1 & 68.3 & 47.5 & 65.4 & 48.4\\
\hline
\small Relation Net\cite{sung2017learning} & 10 & 2 & 65.5 & 40.9 & 62.6 & 41.5\\
\small SoSN \cite{zhang2018power}          & 10 & 2 & 74.4 & 54.6 & 73.0 & 54.2\\
\arrayrulecolor{black} \cdashline{1-7}[1pt/3pt]
\textit{Intra.-Hal.}                       & 10 & 2 & 75.8 & 56.3 & 73.8 & 55.3\\
\textit{Inter.-Hal.}                       & 10 & 2 & 75.6 & 56.4 & 74.2 & 55.6\\
\hline
\small Relation Net\cite{sung2017learning} & 10 & 3 & 69.0 & 45.7 & 67.5 & 46.3\\
\small SoSN \cite{zhang2018power}          & 10 & 3 & 78.0 & 56.3 & 77.5 & 58.6\\
\arrayrulecolor{black} \cdashline{1-7}[1pt/3pt]
\textit{Intra.-Hal.}                       & 10 & 3 & 79.2 & 58.3 & 78.3 & 59.1\\
\textit{Inter.-Hal.}                       & 10 & 3 & 79.3 & 58.5 & 78.6 & 59.9 \\
\hline
\end{tabular}
%}
\vspace{-0.2cm}
\end{table}

\begin{table*}[b]
\vspace{-0.2cm}
\centering
\caption{\small 5-way evaluations on the \textit{mini}Imagenet dataset for different $N$-shot numbers. Refer to \cite{sung2017learning, zhang2018power} for details of baselines.}
\label{table2b}
\makebox[\textwidth]{
\setlength{\tabcolsep}{0.1em}
\renewcommand{\arraystretch}{0.70}
%\fontsize{9}{9}\selectfont
\begin{tabular}{l|c|c|c|c|c|c|c|c|c|c}
\hline
 & \multicolumn{10}{c}{5-way Accuracy} \\ \hline
$W\!$-shot & 1 & 2 & 3 & 4 & 5 & 6 & 7 & 8 & 9 & 10\\ \hline
{\small \textit{Relation Net}} \cite{sung2017learning}& $51.4 \pm\! 0.7$ & $56.7 \pm\! 0.8$ & $60.6 \pm\! 0.8$ & $63.3 \pm\! 0.7$ & $65.6 \pm\! 0.7$  & $66.9 \pm\! 0.7$  & $67.7 \pm\! 0.7$ & $68.6 \pm\! 0.6$ & $69.1 \pm\! 0.6$ & $69.3 \pm\! 0.6$\\
{\small \textit{SoSN}} \cite{zhang2018power}& $53.0 \pm\! 0.8$ & $ 60.8 \pm\! 0.8$ & $ 64.5 \pm\! 0.8$& $ 67.1 \pm\! 0.7$& $ 68.6\pm\! 0.7$ & $70.3\pm\! 0.7$ & $71.5\pm\! 0.6$ & $72.0\pm\! 0.6$ & $72.3\pm\! 0.6$& $73.4\pm\! 0.6$\\
\hline
{\small \textit{w/o Sal. Seg.}} & $ 53.1 \pm\! 0.9$ & $ 60.9 \pm\! 0.8$ & $ 64.7 \pm\! 0.8$& $ 67.3 \pm\! 0.7$& $ 68.9\pm\! 0.7$ & $70.6\pm\! 0.7$ & $71.7\pm\! 0.6$ & $72.1\pm\! 0.6$ & $72.6\pm\! 0.6$& $73.6\pm\! 0.6$\\
{\small \textit{w/o Hal.}} & $ 55.6 \pm\! 0.9$ & $ 63.5 \pm\! 0.8$ & $ 66.2 \pm\! 0.8$& $ 68.2 \pm\! 0.7$& $ 70.4\pm\! 0.7$ & $71.2\pm\! 0.7$ & $72.2\pm\! 0.7$ & $73.2\pm\! 0.6$ & $74.0\pm\! 0.6$& $74.6\pm\! 0.6$\\
\arrayrulecolor{black} \cdashline{1-11}[1pt/3pt]
{\small \textit{Intra.-Hal.}} & $55.6 \pm\! 0.9$ & $63.1 \pm\! 0.8$ & $ 65.9 \pm\! 0.7$ &$ 68.7 \pm\! 0.7$ & $70.8\pm\! 0.7$ & $ 71.8\pm\! 0.7$ & $73.6\pm\! 0.6$ & $73.8\pm\! 0.6$ & $74.1\pm\! 0.6$ & $75.2\pm\! 0.6$\\
{\small \textit{Intra.-Hal.+TriR}} & $55.6 \pm\! 0.9$ & $64.5 \pm\! 0.8$ & $ 67.5 \pm\! 0.7$ &$ 70 .3\pm\! 0.7$ & $71.8\pm\! 0.7$ & $ 72.8\pm\! 0.7$ & $74.1\pm\! 0.6$ & $74.4\pm\! 0.6$ & $74.7\pm\! 0.6$ & $75.7\pm\! 0.6$\\
\arrayrulecolor{black} \cdashline{1-11}[1pt/3pt]
{\small \textit{Inter.-Hal.}} & $ 53.7 \pm\! 0.9$ & $ 58.9 \pm\! 0.8$ & $ 62.4 \pm\! 0.8$& $ 65.2 \pm\! 0.7$& $ 67.7\pm\! 0.7$ & $68.5\pm\! 0.7$ & $69.6\pm\! 0.7$ & $69.9\pm\! 0.6$ & $70.6 \pm\! 0.6$ & $71.1\pm\! 0.6$\\
{\small \textit{Inter.-Hal.+TriR}} & $54.1 \pm\! 0.9$ & $60.1 \pm\! 0.8$ & $ 63.4 \pm\! 0.7$ &$ 65.8 \pm\! 0.7$ & $67.9\pm\! 0.7$ & $ 69.6\pm\! 0.7$ & $70.5\pm\! 0.6$ & $71.0\pm\! 0.7$ & $72.1\pm\! 0.6$ & $72.5\pm\! 0.7$\\
{\small \textit{Inter.-Hal.+TriR+HSP}} & $56.4 \pm\! 0.9$ & $63.0 \pm\! 0.8$ & $ 67.3 \pm\! 0.8$ &$ 69.2\pm\! 0.7$ & $71.0\pm\! 0.6$ & $71.8\pm\! 0.7$ & $72.1\pm\! 0.6$ & $73.0\pm\! 0.6$ & $74.2\pm\! 0.6$ & $75.4\pm\! 0.6$\\
{\small \textit{Inter.-Hal.+TriR+SSP}} & $57.5 \pm\! 0.9$ & $64.8 \pm\! 0.8$ & $ 67.9 \pm\! 0.8$ &$ 70.5\pm\! 0.7$ & $72.0\pm\! 0.7$ & $ 73.2\pm\! 0.7$ & $74.3\pm\! 0.6$ & $74.6\pm\! 0.6$ & $75.2\pm\! 0.6$ & $76.1\pm\! 0.6$\\
\hline
\end{tabular}}
\vspace{-0.2cm}
\end{table*}
%\fi

\vspace{-0.1cm}
\subsection{Results}
\label{sec:resres}
For \textit{mini}Imagenet dataset, Table \ref{table2} shows that our proposed SalNet outperforms all other state-of-the-art methods on standard 5-way 1- and 5-shot protocols. Compared with current state-of-the-art methods, our {\em SalNet Inter-class Hal.} model achieves $\sim$4.4\% and $\sim$3.3\% higher top-1 accuracy than SoSN on 1- and 5-shot protocols,  respectively, while our {\em SalNet Intra-class Hal.} yields improvements of $\sim$2.5\% and $\sim$3.1\% accuracy over SoSN.

\begin{figure}[t]
\vspace{-0.3cm}
\centering
\includegraphics[width=\linewidth]{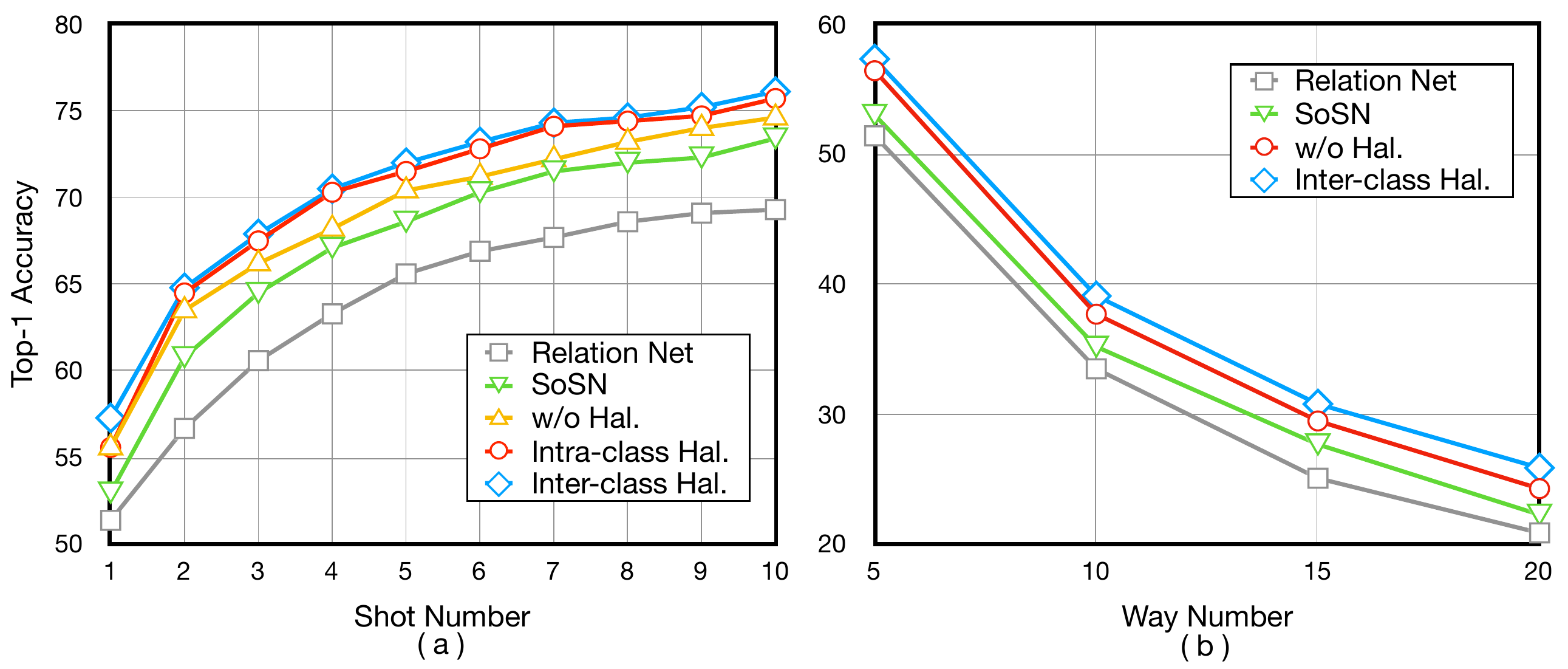}
\vspace{-0.7cm}
\caption{\small The accuracy as a function of ({\em left}) $W\!$-shot (5-way) and ({\em right}) $N$-way (5-shot) numbers on \textit{mini}Imagenet given different methods. Our models improve results over all baselines.}
\label{fig:plot1}
\vspace{-0.1cm}
\end{figure}

\begin{figure}[t]
\centering
\includegraphics[width=\linewidth]{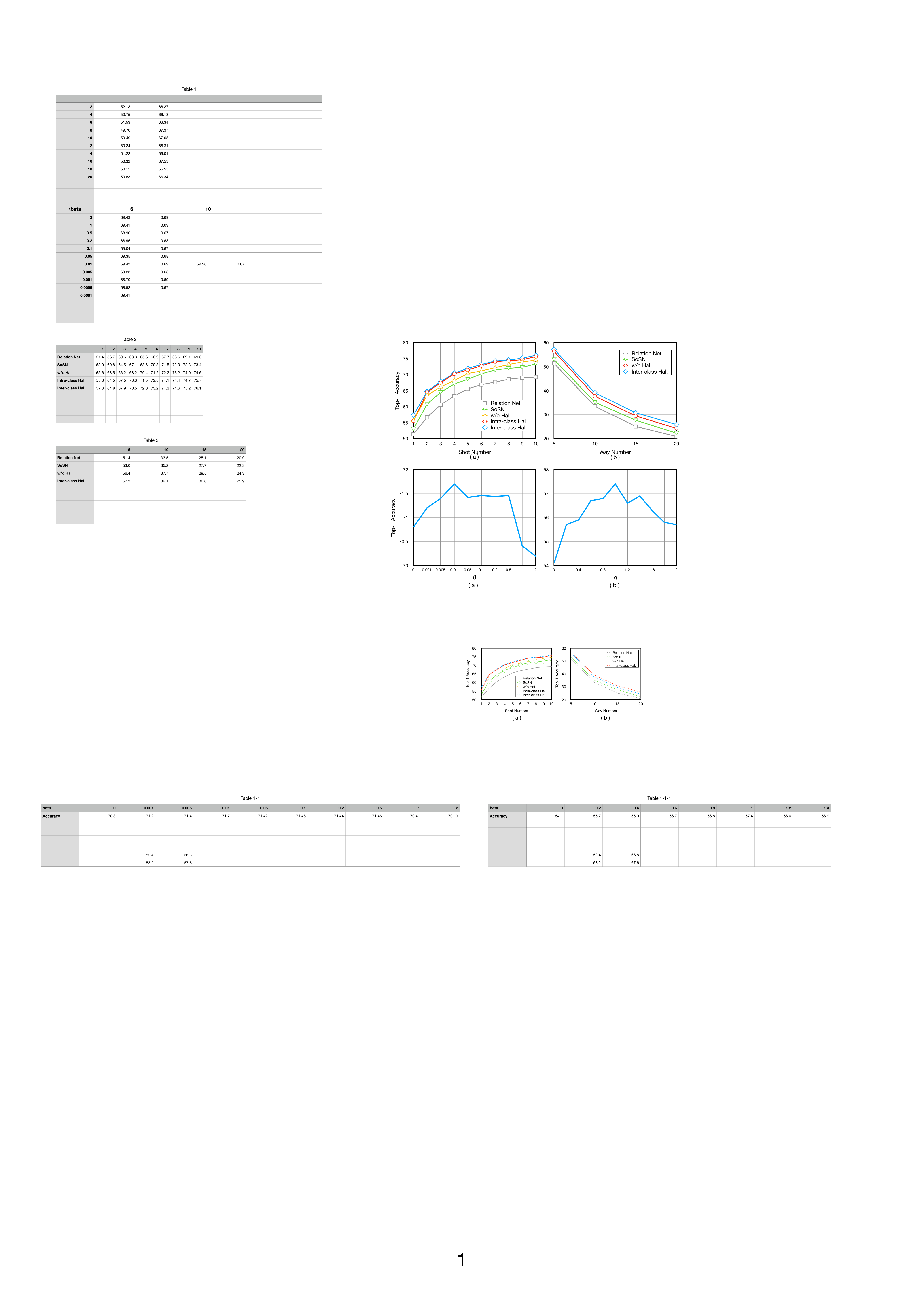}
\vspace{-0.7cm}
\caption{\small The accuracy  on \textit{mini}Imagenet as a function of (a) $\beta$ of TriR from Eq. \eqref{eq:trir} (5-shot 5-way) and (b) $\alpha$ of SSP from Eq. \eqref{eq:ssp} (1-shot 5-way).}
\label{fig:plot2}
\vspace{-0.3cm}
\end{figure}

Table \ref{table5} presents results on Open MIC. The improvements of {\em SalNet Inter-class Hal.} and {\em SalNet Intra-class Hal.} on this dataset are consistent with \textit{mini}Imagenet. However, the improvements on some splits are small (\ie, $\sim$ 1.1\%) due to the difficulty of these splits \eg, jewellery, fossils, complex non-local engine installations or semi-transparent exhibits captured with wearable cameras %for Open MIC
cannot be easily segmented out by saliency detectors.

\noindent\textbf{Ablation study.} The network proposed in our paper builds on the baseline framework \cite{sung2017learning}. However, we have added several non-trivial units/sub-networks to accomplish our goal of the datapoint hallucination in the feature space. Thus, we perform additional experiments to show that the achieved accuracy gains  stem from our contributions. We also break down the accuracy w.r.t. various components. 

Firstly, Table \ref{table2} shows that if the saliency segmentation and data hallucination are disabled in our pipeline ({\em SalNet w/o Sal. Seg.}), the performance on all protocols drops down to the baseline level of SoSN. %, which demonstrates that the benefits from network modification are very limited. 

Moreover, we observe that SalNet outperforms SoSN even if we segment images into foregrounds and backgrounds and pass them via our network  without the use of hallucinated datapoints ({\em SalNet w/o Hal.}). We assert that such improvements stem from the ability of the saliency detector to localize main objects in images. This is a form of spatial knowledge transfer which helps our network capture the similarity between query and support images better. 

Figure \ref{fig:plot1} (a) shows the accuracy of our ({\em SalNet Intra-class Hal.}) model on \textit{mini}Imagenet for 5-shot 5-way case as a function of the parameter $\beta$ of our regularization loss TriR. We observe that for $\beta\!=\!0.01$ we gain $\sim$1\% accuracy over $\beta\!=\!0$ (TriR disabled). Importantly, the gain remains stable over a large range $0.005\!\leq\!\beta\!\leq\!0.5$.
Table \ref{table2b} verifies further the usefulness of our TriR regularization in combination with the intra- and inter-class hallucination SalNet ({\em  Intra.-Hal.+TriR}) and ({\em Inter.-Hal.+TriR}) with gains up to $1.6\%$ and $1.5\%$ accuracy on \textit{mini}Imagenet. We conclude that TriR helps our end-to-end training by forcing encoder $\mathbf{g}$ to mimic teacher $\mathbf{g}^*$ for real foreground-background pairs ($\mathbf{g}^*$ is trained on such pairs only to act as a reliable superv.).

Figure  \ref{fig:plot1}  (b) shows the accuracy of our ({\em SalNet Inter-class Hal.}) model on \textit{mini}Imagenet for 1-shot 5-way  as a function of the Soft Similarity Prior (SSP). The maximum observed gain in accuracy is $\sim$3.3\%.
Table \ref{table2b} further compares the hard and soft priors ({\em SalNet Inter-class Hal.+HSP}) and ({\em SalNet Inter-class Hal.+SSP}) with SSP outperforming HSP by up to $\sim$2.2\%.

\begin{figure}[t]
\vspace{-0.3cm}
\centering
\includegraphics[width=\linewidth]{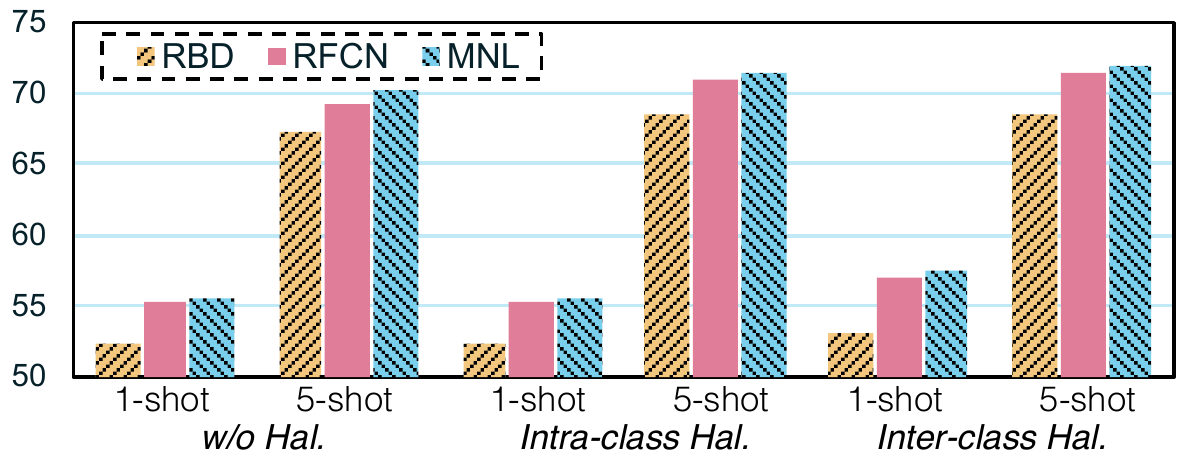}
\caption{\small The results on RBD \cite{Background-Detection:CVPR-2014}, RFCN \cite{RFCN} and MNL \cite{Zhang_2018_CVPR} saliency methods for \textit{mini}Imagenet.} \label{fig:saliency_comp}
\vspace{-0.3cm}
\end{figure}

Lastly, Figure \ref{fig:saliency_comp} compares several saliency methods in terms of few-shot learning accuracy. The complex saliency methods perform equally well. However, the use of the RBD approach \cite{Background-Detection:CVPR-2014} results in a significant performance loss due to its numerous failures \eg, see Figure \ref{fig:sal_map}.

%\noindent{\textbf{ Computational complexity.}} Evaluating one saliency map with RBD, RFCN  and MNL takes 80, 300 and 250 ms, respectively. Training SalNet, SoSN and Relation Net over 100 episodes takes 26, 22, 12 seconds. 

\begin{figure}[t]
    \vspace{-0.3cm}
    \centering
    \includegraphics[width=6cm]{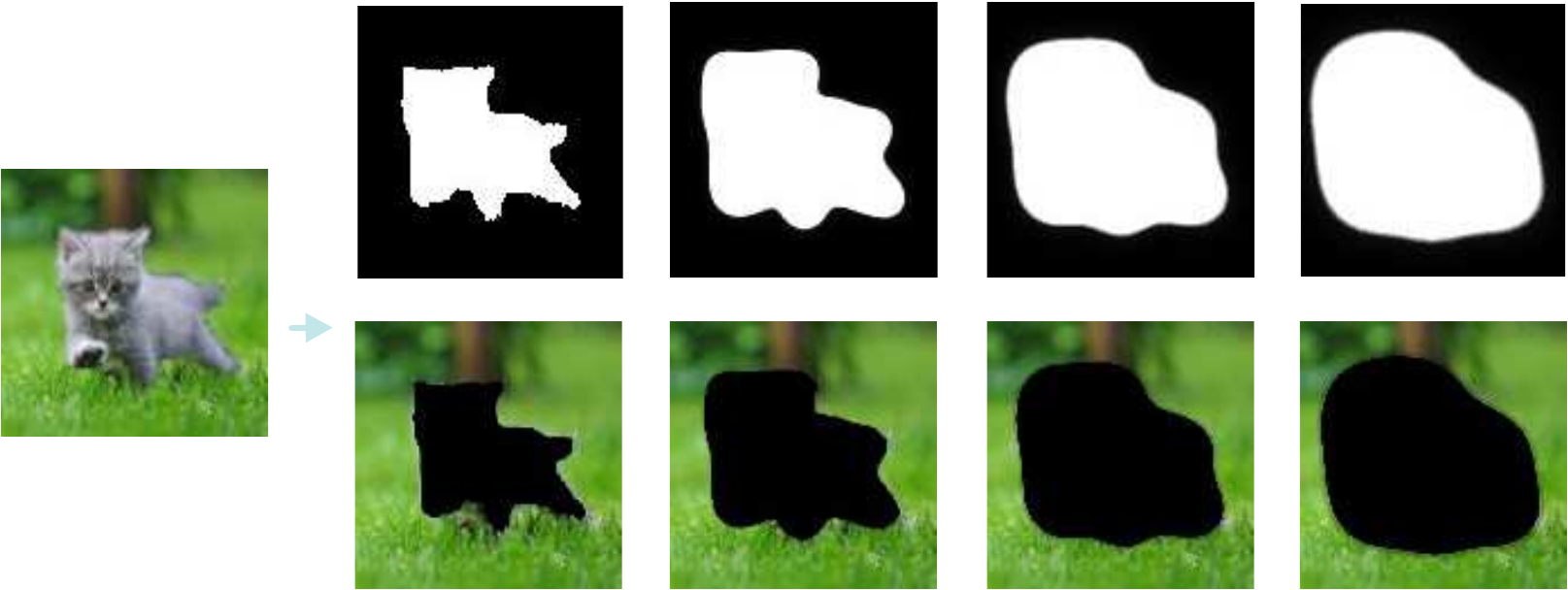}
		\vspace{-0.1cm}
    \caption{\small Gradual dilation of the foreground mask. $\!\!\!\!$}
    \label{fig:cats}
    \vspace{-0.1cm}
\end{figure}

\noindent\textbf{Saliency Map Dilation.}
As backgrounds extracted via a saliency detector contain `cut out' silhouettes, they unintentionally carry some foreground information. Figure \ref{fig:cats} suggests that if we apply the Gaussian blur and a threshold over the masks to eliminate the silhouette shapes, we can prevent mixing the primary foreground with a foreground corresponding to silhouettes. Table \ref{table-sil} shows that pairing each foreground with background images whose silhouettes were removed by dilating according to two different radii ({\em Dilation}) leads to further improvements due to doubling of possible within-class combinations for ({\em Intra-class Hal.}).

%We investigate the dilation strategy on our baseline mixing network (w/o hallucination). It can be seen in Figure \ref{fig:sal_blurry} that applying dilation for background extraction can further improve the performance.

\comment
{
\begin{figure}[t]
    \centering
    \includegraphics[width=0.9\linewidth]{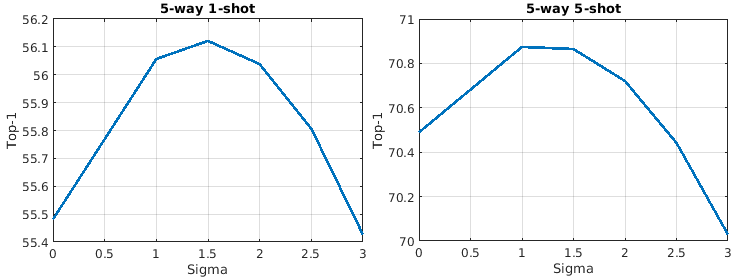}
		\vspace{-0.1cm}
    \caption{\small Top-1 accuracy on \textit{mini}imagenet. ($\sigma\!=\!1-3$).$\!\!\!\!$}
    \label{fig:sal_blurry}
    \vspace{-0.3cm}
\end{figure}
}
%\fi

\begin{table}[t]
%\vspace{-0.3cm}
%
%\makebox[\linewidthwidth]{
\setlength{\tabcolsep}{0.10em}
\renewcommand{\arraystretch}{0.70}
%\fontsize{7.5}{9}\selectfont
\begin{tabular}{l|c|c} 
\hline
%Model  & \multicolumn{2}{c}{5-way Acc.} \\ 
 Model & 5-way 1-shot & 5-way 5-shot \\ \hline
\textit{{\small Intra-class Hal.}}  & $55.57 \pm 0.86$ & $ 71.78\pm 0.69$ \\
\textit{{\small Intra-class Hal.+Dilation}} & $\mathbf{56.67} \pm 0.85$ & $ \mathbf{72.15}\pm 0.68$ \\
%\textit{{\small Inter-class Hal.}} & $ 57.45 \pm 0.88$ & $ 72.01\pm 0.67$ \\
%\textit{{\small Inter-class Hal.+Dilation}}  & $ \mathbf{57.96} \pm 0.87$ & $ \mathbf{72.61}\pm 0.65$ \\
\hline
\end{tabular}
%}
\centering
\caption{\small Results for dilating contours of silhouettes.}
\label{table-sil}
\vspace{-0.3cm}
\end{table}
\vspace{-0.2cm}
\section{Conclusions}
In this paper, we have presented two novel light-weight data hallucination strategies for few-shot learning. in contrast to other costly hallucination methods based on GANs, we have leveraged the readily available saliency network to obtain foreground-background pairs on which we trained our SalNet network in end-to-end manner. To cope with noises of saliency maps,  we have proposed a Real Representation Regularization (TriR) which regularizes our network with viable solutions for real foreground-background pairs. To alleviate performance loss caused by implausible foreground-background hypotheses, we have proposed a similarity-based priors  effectively reduced the influence of incorrect hypotheses. For future work, we will investigate a self-supervised attention module for similarity perception and study relaxations of saliency segmentation methods. %Kindly see our supp. material for additional results.

%\small
{
\noindent
\textbf{Acknowledgements.} This research is supported by the China Scholarship Council (CSC Student ID 201603170283). We also thank CSIRO Scientific Computing, NVIDIA (GPU grant) and National University of Defense Technology for their support.
}

\begin{appendices}
\section{Saliency Maps on the Open MIC dataset}

\begin{figure}[h]
	\vspace{-0.2cm}
	\centering
	\includegraphics[width=\linewidth]{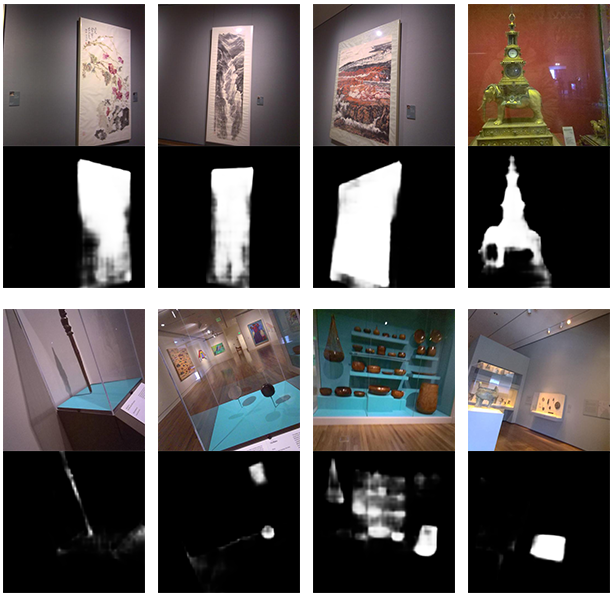}
	\caption{Examples of saliency maps on the Open MIC dataset. The MNL detector was used.}
	\label{fig:openmic_sal}
	\vspace{-0.2cm}
\end{figure}

In Figure \ref{fig:openmic_sal}, we present saliency maps for some exhibit instances from the Open MIC dataset. Many exhibits can be filtered out reliably. However, saliency maps for composite scenes containing numerous exhibits are the ones most likely to fail. In the future, we will investigate how to improve the use of  such unreliable saliency maps for such exhibits. Note that our results on exhibitions containing such composite scences still benefit from our approach--our mixing network can reduce the noise from saliency maps.

\section{Evaluations for $\mathbf{224\!\times\!224}$ pixel images}
\begin{figure*}[h]
	\vspace{-0.6cm}
	\centering
	\includegraphics[width=\linewidth]{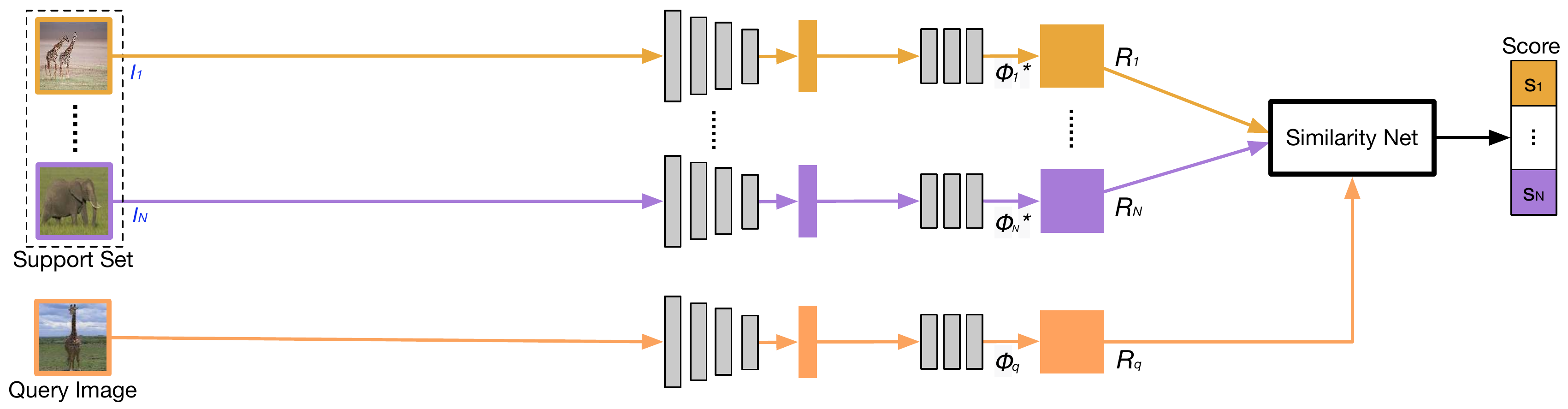}
	\caption{The network architecture of {\bf baseline 1} 'w/o Sal. Seg.'. It can be seen that once the Saliency Net and data hallucination strategies are disabled, the network pipeline are very similar to SoSN . Note that we write $\mathbf{R}_1,\cdots,\mathbf{R}_N$ for brevity rather than $\mathbf{R}_{11},\cdots,\mathbf{R}_{NN}$ (as dictated by Eq. (4) of our main submission) as no hallucination takes place here \eg, we evaluate only $\mathbf{R}_{ij}$ for $i\!=\!j$. Moreover, note that $\Phi_i$ are not generated by the foreground-background mechanism from Eq. (3) of our main submission. Instead, entire images are encoded.}
	\label{fig:baseline1}
	%\vspace{-0.2cm}
\end{figure*}

We employ $84\!\times\!84$ image in our experiments for fair comparison with other state-of-the-art methods presented in our paper. However, it is easy to use large size images in our network without its modifications due to the ability of second-order representations to aggregate variable number of feature vectors into a fixed-size matrix (our relationship descriptors are stacked matrices). Here we apply $224\times224$ image to demonstrate the benefits from larger image size. 

\begin{table}[h]
\vspace{-0.3cm}
\centering
\caption{\small Accuracy on the \textit{mini}Imagenet dataset given different size of images. See \cite{sung2017learning, zhang2018power} for details of baselines. The astersik ({\em *}) denotes the `sanity check' results on our proposed pipeline given disabled both saliency segmentation and hallucination.}
\label{table2}
%\makebox[\linewidthwidth]{
\setlength{\tabcolsep}{0.10em}
\renewcommand{\arraystretch}{0.70}
%\fontsize{8.5}{9}\selectfont
\begin{tabular}{l|c|c|c} 
\hline
Model & Fine & \multicolumn{2}{c}{5-way Acc.} \\ 
& Tune & 1-shot & 5-shot \\ \hline
\textit{Matching Nets} \cite{vinyals2016matching}& N & \,\,$43.56 \pm 0.84$\,\, & \,\,$55.31 \pm 0.73$\,\,  \\
\textit{Meta Nets} \cite{munkhdalai2017meta} & N & $49.21 \pm 0.96$ & - \\
\textit{Meta-Learn Nets} \cite{ravi2016optimization} & N & $43.44 \pm 0.77$ & $60.60 \pm 0.71$ \\
\textit{Prototypical Net} \cite{snell2017prototypical}& N & $49.42 \pm 0.78$ & $68.20 \pm 0.66$ \\ 
\textit{MAML} \cite{finn2017model}& Y & $48.70 \pm 1.84$ & $63.11 \pm 0.92$ \\ 
\textit{Relation Net}  \cite{sung2017learning}& N & $51.36 \pm 0.86$ & $65.63 \pm 0.72$  \\
\textit{SoSN} \cite{zhang2018power}& N & $52.96 \pm 0.83$ & $68.63 \pm 0.68$ \\
\hline
\multicolumn{4}{c}{84$\times$84} \\ \hline
\textit{SalNet {\small w/o Sal. Seg.}} (*) & N & $ 53.15\pm 0.87$ & $ 68.87 \pm 0.67$ \\
\textit{SalNet {\small w/o Hal.}} & N & \multirow{2}{*}{$55.57 \pm 0.86$} & $70.35 \pm 0.66$ \\
\textit{SalNet {\small Intra-class Hal.}} & N &  & $ 71.78\pm 0.69$ \\
%\textit{SalNet {\small Interclass-Hal.}} & N & $ 54.12 \pm 0.83\%$ & $ 67.65\pm 0.68\%$ \\
\textit{SalNet {\small Inter-class Hal.}} & N & $ 57.45 \pm 0.88$ & $ 72.01\pm 0.67$ \\
\hline
\multicolumn{4}{c}{224$\times$224} \\ \hline
SoSN\cite{zhang2018power} & N & $ 59.22\pm 0.91$ & $ 73.24 \pm 0.69$ \\
\textit{SalNet {\small w/o Sal. Seg.}} (*) & N & $ 60.36\pm 0.86$ & $ 74.34 \pm 0.67$ \\
\textit{SalNet {\small w/o Hal.}} & N & \multirow{2}{*}{$62.22 \pm 0.87$} & $ 76.86 \pm 0.65$ \\
\textit{SalNet {\small Intra-class Hal.}} & N &  & $ 77.95\pm 0.65$ \\
\textit{SalNet {\small Inter-class Hal.}} & N & $ 63.88 \pm 0.86$ & $ 78.34\pm 0.63$ \\
\hline
\end{tabular}
\vspace{-0.1cm}
\end{table}

\section{Network Architecture of Our Baseline Models and Additional Experiments for TriR}

\begin{figure*}[h]
	\vspace{-0cm}
	\centering
	\includegraphics[width=\linewidth]{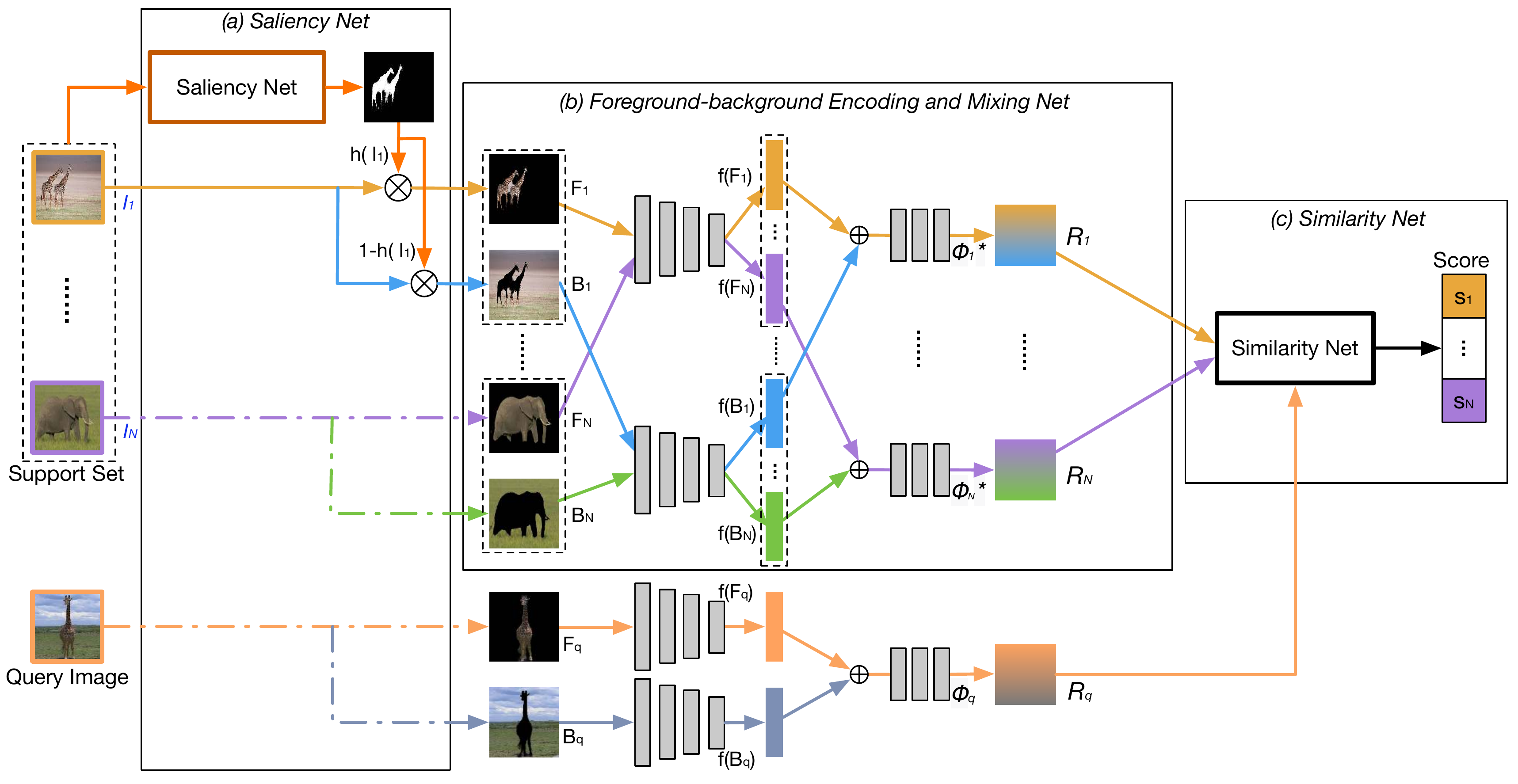}
	\caption{The network architecture of {\bf baseline 2} 'w/o Hallucination'. Note that although the data hallucination mechanism is disabled, we still apply saliency maps to segment foregrounds and backgrounds as we want the TriR loss to learn to account for the potential noise stemming from the foreground-background segmentation which is used in the main network. Moreover, we write $\mathbf{R}_1,\cdots,\mathbf{R}_N$ for brevity rather than $\mathbf{R}_{11},\cdots,\mathbf{R}_{NN}$ (as dictated by Eq. (4) of our main submission) as no hallucination takes place here \eg, we evaluate only $\mathbf{R}_{ij}$ for $i\!=\!j$. Note that $\Phi_i$ are generated by the foreground-background mechanism in Eq. (3) of our main paper (we abbreviate $\Phi_{ii}$ to $\Phi_i$).}
	\label{fig:baseline2}
	\vspace{-0.2cm}
\end{figure*}

\begin{table}[t]
\centering
\caption{\small Evaluations on the \textit{mini}Imagenet dataset given different teacher networks for the TriR regularization.}
\label{table2}
\vspace{-0.2cm}
%\makebox[\linewidthwidth]{
\setlength{\tabcolsep}{0.30em}
\renewcommand{\arraystretch}{0.75}
%\fontsize{8.5}{9}\selectfont
\begin{tabular}{l|c|c|c} 
\hline
Model & Fine & \multicolumn{2}{c}{5-way Acc.} \\ 
& Tune & 1-shot & 5-shot \\ \hline
\multicolumn{4}{c}{{\bf baseline 1} (opt. (i)) as a teacher network in TriR} \\ \hline
\textit{SalNet {\small Intra-class Hal.}} & N & $ 56.11\pm 0.88$ & $ 71.56\pm 0.67$ \\
\textit{SalNet {\small Inter-class Hal.}} & N & $ 57.24 \pm 0.94$ & $ 72.49\pm 0.65$ \\
\hline
\multicolumn{4}{c}{{\bf baseline 2} (opt. (ii)) as a teacher network in TriR} \\ \hline
\textit{SalNet {\small Intra-class Hal.}} & N & $55.57 \pm 0.86$ & $ 71.78\pm 0.69$ \\
\textit{SalNet {\small Inter-class Hal.}} & N & $ 57.45 \pm 0.88$ & $ 72.01\pm 0.67$ \\
\hline
\end{tabular}
\end{table}

Below we present the diagrams of two baseline networks used in our paper. The {\bf baseline 1} in Figure \ref{fig:baseline1} is the original pipeline 'w/o Sal. Seg.', which is trained without saliency segmentation or data hallucination -- it is very similar to the SoSN pipeline \cite{zhang2018power}. Figure \ref{fig:baseline2} demonstrates the {\bf baseline 2} 'w/o Hal.', which employs saliency network to segment the foregrounds and backgrounds but does not hallucinate the data (no mixing of a foreground with numerous different backgrounds is allowed).

In our paper, the reported results are obtained by using {\bf baseline 2} 'w/o Hal.' pipeline as teacher in TriR regularization (option (ii) in line 513 of our main submission). However, for completeness, we also investigate {\bf baseline 1} 'w/o Hal.' pipeline as teacher in TriR regularization (option (i) in line 512 of our main submission). Table \ref{table2} shows that both TriR teachers perform similarly to each other.

\end{appendices}
{\small
\bibliographystyle{ieee_fullname}
\bibliography{fsl}
}

\end{document}